\documentclass{article} 
\usepackage{colm2024_conference}
\usepackage{pdfpages}
\usepackage{microtype}
\usepackage{hyperref}
\usepackage{url}
\usepackage{amsmath}
\usepackage{booktabs}
\usepackage{natbib}
\definecolor{darkblue}{rgb}{0, 0, 0.5}
\hypersetup{colorlinks=true, citecolor=darkblue, linkcolor=darkblue, urlcolor=darkblue}
\usepackage{graphicx}       
\usepackage{subcaption}     
\usepackage[most]{tcolorbox}
\usepackage{makecell}       
\usepackage{tabularx}
\usepackage{ulem}

\title{Writing-Zero: Bridge the Gap Between Non-verifiable Tasks and Verifiable Rewards}
\author{ Ruipeng Jia, Yunyi Yang, Yongbo Gai, Kai Luo, \\ Shihao Huang, Jianhe Lin, Xiaoxi Jiang, Guanjun Jiang \\
 \vspace{0.3cm}
 \textbf{Quark LLM, Alibaba Group} \\
}

\begin{document}

\maketitle

\begin{abstract}

Reinforcement learning with verifiable rewards (RLVR) has facilitated significant advances in large language models (LLMs), 
particularly for reasoning tasks with objective, ground-truth answers, such as math and code generation. 
However, a substantial gap persists for non-verifiable tasks—such as creative writing and open-ended dialogue—where 
quality assessment is inherently subjective and lacks definitive, externally verifiable references. 
Existing methodologies for these tasks predominantly rely on scalar reward models trained using human preferences, 
but these models suffer from limited generalization and are vulnerable to reward hacking, 
including issues such as over-explanation and length bias. 
In this work, we propose a unified RLVR-based training paradigm that effectively bridges the gap 
between non-verifiable tasks and verifiable reward. 
We introduce a novel pairwise Generative Reward Model (GenRM) grounded in writing principles 
and a new Bootstrapped Relative Policy Optimization (BRPO) algorithm. 
The pairwise writing GenRM applies self-principled critique to transform subjective assessments into robust, verifiable rewards, 
while BRPO facilitates dynamic, reference-free pairwise comparisons by utilizing bootstrapped responses 
as temporary references during group rollouts in reinforcement learning (RL) training. 
Our approach enables LLMs to cultivate advanced writing capabilities without requiring supervised fine-tuning. 
This is demonstrated by Writing-Zero, which exhibits consistent performance improvements 
and enhanced resilience to reward hacking compared to scalar reward baselines. 
In addition, our method achieves competitive results on both proprietary and publicly available writing benchmarks. 
These results suggest the potential for unifying rule-based, reference-based, and reference-free reward modeling 
within the RLVR framework, thereby advancing the development of a comprehensive and scalable RL training paradigm 
with broad applicability across various language tasks.
\end{abstract}

\section{Introduction}

Recently, large language models (LLMs) have achieved remarkable breakthroughs in reasoning capabilities through reinforcement learning with verifiable rewards (RLVR),
particularly in solving complex logical tasks such as mathematics and programming \citep{guo2025deepseek}.
RLVR relies on reference-based signals, where the availability of objective ground-truth answers enables reliable verification of model responses.
This approach has proven particularly effective in tasks with well-defined solutions,
such as mathematical reasoning and code generation, where simple rule-based verifiers can provide clear binary signals (correct or incorrect) \citep{team2025kimi}.
However, there is a spectrum ranging from verifiable to non-verifiable problems,
with problems like mathematics and coding at one end,
multi-subject QA with less structured answers in the middle,
and creative writing, multi-turn dialogue that lack a reference answer and require quality assessment based on human preferences at the non-verifiable end.
For less-verifiable or non-verifiable problems, previous works mainly rely on a scalar reward model trained with human preference data for RLHF training,
which has limited generalization ability and is prone to reward hacking \citep{zhong2025comprehensive}.
For example, in creative writing scenarios, models trained with RLHF often exhibit over-explanation,
where they append lengthy justifications of how their response perfectly meets user requirements,
even when the actual content fails to do so.

\begin{figure}[htbp]
    \centering
    \includegraphics[width=0.8\textwidth]{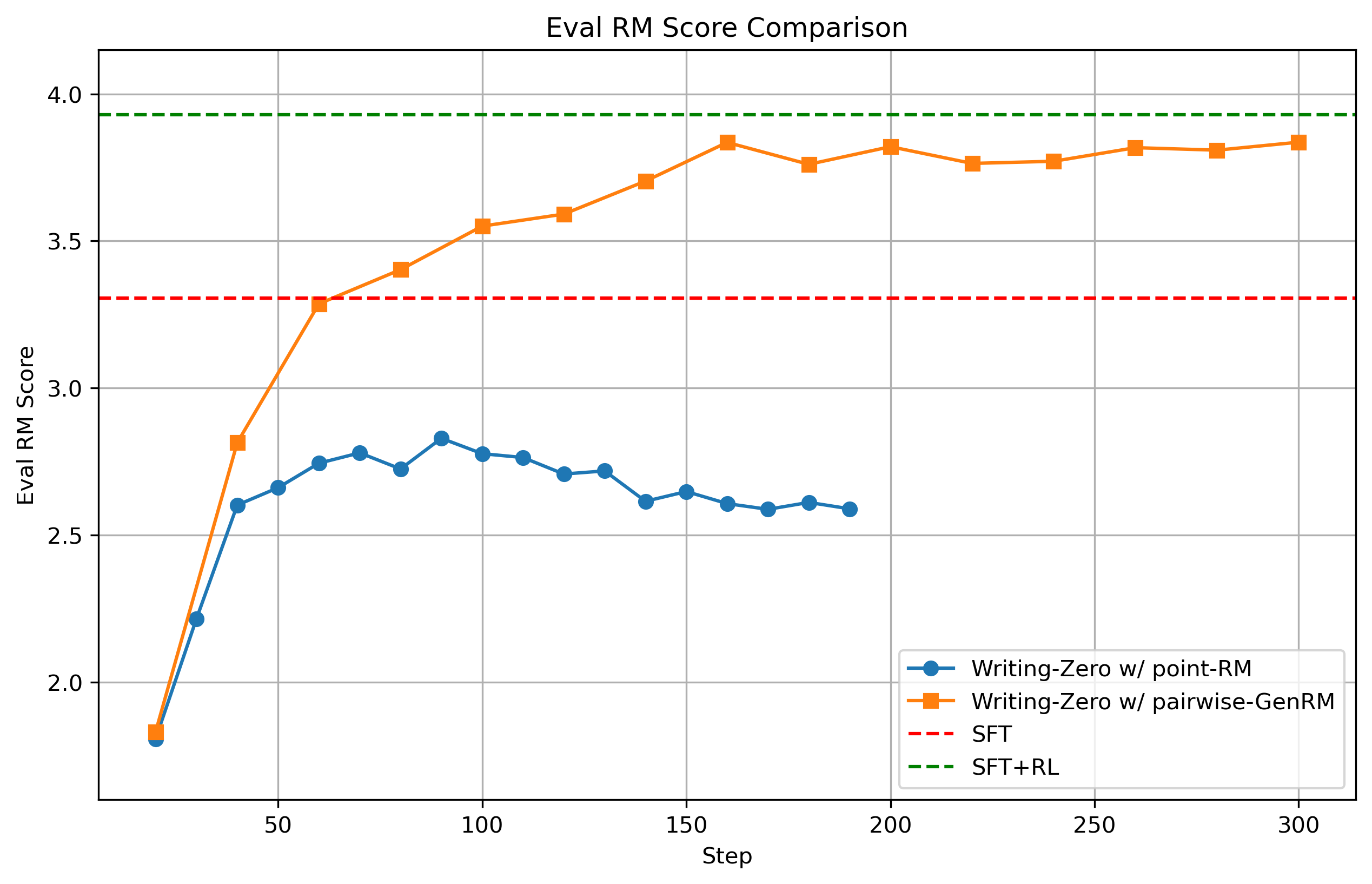}
                                                \caption{Comparison of Eval RM scores during \textbf{Writing-Zero} training.
        The blue line shows \texttt{Qwen3-32B-Base-ScalarRM-GRPO},
        while the orange line shows \textbf{Writing-Zero} (\texttt{Qwen3-32B-Base-GenRM-BRPO}).
        The dashed red and green lines indicate the SFT (Writing-SFT) and SFT+RL (Writing-SFT-GenRM-BRPO) baselines, respectively.}
    \label{fig:writing_zero}
\end{figure}

Developing high-quality and robust reward modeling methods for more general domains remains a challenging and active research direction.
\cite{su2025expanding} propose training a generative reward model to verify whether the response matches the reference answer,
expanding the application of RLVR to less structured domains such as multi-subject QA.
Most recently, researchers attempt directly adapting RLVR to improve generative reward modeling (GenRM)
and train GRM in R1-style with long COT of designed evaluation principles and critiques,
which results in more accurate and reliable rewards across broader domains and effective test-time scaling \citep{liu2025inference, chen2025rm}.
While the emergence of reliable and robust GenRM has opened promising avenues for extending RLVR to broader domains,
its effective application, particularly to non-verifiable tasks, remains a challenging and largely unexplored research topic.

As the Chinese saying goes, there is no absolute first place in literary arts,
creative writing represents one of the typical non-verifiable tasks where quality assessment is inherently subjective and lacks definitive reference.
In this work, we aim to bridge the gap between the non-verifiable tasks and verifiable rewards and propose a new training paradigm for non-verifiable writing tasks,
with writing-principle-based pairwise GenRM and a new RL algorithm Bootstrapped Relative Policy Optimization (BRPO).
Specifically, inspired by \cite{liu2025inference}, we first train the Pairwise Writing GenRM through high-quality self-principled critique cold-start data,
and perform RLVR to refine the GenRM's ability to formulate adaptive principles and nuanced critiques specific to different writing scenarios and diverse response pairs,
thereby producing more reliable outcome rewards for creative writing evaluation.
The Writing-GenRM takes in a pair of responses and outputs two respective scores between 0 and 10 to indicate comparative quality,
effectively transforming subjective assessments into reliable verifiable rewards.
Then we introduce Bootstrapped Relative Policy Optimization (BRPO),
which dynamically selects samples from within the group rollouts as temporary references for pairwise comparison and advantage estimation.
This bootstrap mechanism eliminates the need for fixed external references,
enabling the policy model to continuously improve by leveraging its own increasingly sophisticated outputs for comparison.
Similar to DeepSeek-R1 \citep{guo2025deepseek}, we explore the potential of LLMs developing writing capabilities without any supervised finetuning data
and introduce Writing-Zero using Qwen3-32B-Base \citep{yang2025qwen3} as the base model.
Throughout the training process, Writing-Zero shows consistent improvement and exceptional resistance against reward hacking
compared with the base model trained with scalar reward, as shown in \figurename~\ref{fig:writing_zero}.
Furthermore, we introduce Writing-R1 using an in-house thinking SFT model as the base model and achieve competitive results on in-house and open-source writing benchmarks.
Our empirical results demonstrate the effectiveness of our proposed method.

To sum up, our work represents an important step in unifying different reward modeling paradigms under the RLVR training framework.
We demonstrate that by leveraging pairwise generative reward modeling with self-principled critique,
even non-verifiable tasks like creative writing can benefit from stable and scalable RL training, similar to verifiable tasks.
This opens up the possibility of unifying three major reward modeling approaches: rule-based rewards for well-defined tasks,
model-based rewards with reference answers for less-structured tasks, and model-based rewards without reference answers for creative tasks.
Our work thus paves the way for establishing a comprehensive and consistent RLVR training paradigm that can be applied across the entire spectrum of language tasks,
from verifiable to non-verifiable domains.

\section{Preliminary}

\subsection{Group Relative Policy Optimization}

GRPO \citep{shao2024deepseekmath} is an efficient and effective RL algorithm, by introducing the group-relative normalization to estimate the advantage and eliminating the value function of PPO \citep{schulman2017proximal}.
For a specific question $q$ in dataset $\mathcal{D}$, the behavior policy $\pi_{\theta_{old}}$ samples a group of $G$ individual responses $\{o_i\}_{i=1}^G$.
Then, the advantage of the $i$-th response is calculated by normalizing the group-level rewards $\{R_i\}_{i=1}^G$.
Like PPO, GRPO also stabilizes training and improves sample efficiency by constraining the policy model $\pi_{\theta}$ updates within a proximal region with $\text{clip}()$.
Specifically, GRPO updates the policy by maximizing the following objective:
\begin{equation}
  \begin{aligned}
    \mathcal{J}_{\text{GRPO}}(\theta) & = \mathbb{E}_{q \sim \mathcal{D}, \{o_i\}_{i=1}^G \sim \pi_{\theta_{\text{old}}}(\cdot|q)} \\
                                      & \left[\frac{1}{G} \sum_{i=1}^{G} \frac{1}{|o_i|} \sum_{t=1}^{|o_i|} \left( \min \left( r_{i,t}(\theta) \hat{A}_{i,t}, \ \text{clip} \left( r_{i,t}(\theta), 1-\varepsilon, 1+\varepsilon \right) \hat{A}_{i,t} \right) - \beta D_{\text{KL}}(\pi_{\theta} \| \pi_{\text{ref}}) \right) \right]
  \end{aligned}
\end{equation}

\begin{equation}
\label{grpo_advantage}
r_{i,t}(\theta) = \frac{\pi_{\theta}(o_{i,t} \mid q, o_{i,<t})}{\pi_{\theta_{\text{old}}}(o_{i,t} \mid q, o_{i,<t})} \qquad
\hat{A}_{i,t} = \frac{R_i - \text{mean}(\{R_i\}_{i=1}^G)}{\text{std}(\{R_i\}_{i=1}^G)}
\end{equation}

where $\hat{A}_t$ is the estimator of the advantage at time step $t$, $\varepsilon$ is the clipping range of importance sampling ratio $r_{i,t}$, and $\pi_{\text{ref}}$ is the reference model, same as the initial $\pi_{\theta}$.

\subsection{Dynamic Sampling Policy Optimization}
Compared to GRPO, DAPO \citep{yu2025dapo} introduces another four important tricks: clip-higher, dynamic sampling, token-level policy gradient loss and overlong reward shaping.
The key improvement is dynamic sampling, by over-sampling and filtering out prompts with the accuracy equal to 1 and 0, leaving all prompts in the batch with effective gradients,
avoiding dampening the gradient signals for model training with larger variance in gradient.
Specifically, with question-answer pairs $(q, a)$ in dataset $\mathcal{D}$, DAPO updates the policy by maximizing the following objective:

\begin{equation}
  \begin{aligned}
  \mathcal{J}_{\text{DAPO}}(\theta) =& \ \ \mathbb{E}_{(q,a) \sim \mathcal{D}, \{o_i\}_{i=1}^G \sim \pi_{\theta_{\text{old}}}(\cdot|q)} \\
  & \left[ \frac{1}{\sum_{i=1}^G |o_i|} \sum_{i=1}^G \sum_{t=1}^{|o_i|} \min \left( r_{i,t}(\theta) \hat{A}_{i,t}, \ \text{clip} \left( r_{i,t}(\theta), 1-\varepsilon_{\text{low}}, 1+\varepsilon_{\text{high}} \right) \hat{A}_{i,t} \right) \right] \\
  \text{s.t.} \ \ & 0 < \left| \{o_i \mid \text{is\_equivalent}(a, o_i)\} \right| < G,
  \end{aligned}
\end{equation}

\subsection{Rule-based Reward Modeling}
For verifiable tasks, defining a rule-based reward function is straightforward. This approach yields accurate reward signals, which are crucial for scaling the model's reasoning capability through RL, while mitigating issues like reward hacking. With $y$ as the ground-truth answer and $\hat{y}$ as the predicted answer, such a reward function can be simply defined as

\begin{equation}
R(\hat{y}, y) =
\begin{cases}
1, & \text{is\_equivalent}(\hat{y}, y) \\
-1, & \text{otherwise}
\end{cases}
\end{equation}

For non-verifiable tasks, mainstream RLHF approaches still largely employ scalar Bradley-Terry reward models.

\section{Approach}

In this paper, we propose a new training paradigm for the non-verifiable writing tasks via Reinforcement Learning with Verifiable Rewards (RLVR),
which consists of two stages: first, we train a Self-Principled Critique Pairwise GenRM for Creative Writing;
second, we train writing models with the Pairwise Writing GenRM through Bootstrapped Relative Policy Optimization (BRPO).

\subsection{Self-Principled Critique Pairwise GenRM for Creative Writing}
\label{writing_genrm}

The Self-Principled Critique Tuning (SPCT) framework, proposed by \cite{liu2025inference}, offers a method for unified scoring of single or multiple responses via textual critique.
Building upon this, we introduce our \textbf{Pairwise Writing GenRM}, which adapts and refines the SPCT methodology specifically for the nuances of creative writing. Our approach incorporates two key modifications to enhance its suitability and effectiveness for non-verifiable writing tasks:

\begin{itemize}
    \item \textbf{Focused Application on Non-verifiable Writing:} Unlike the broader application of SPCT, our Pairwise Writing GenRM is exclusively applied to non-verifiable writing tasks. This focus stems from the observation that verifiable tasks, which possess definitive answers, can be more directly and reliably assessed using rule-based verifiable rewards.
    \item \textbf{Dedicated Pairwise Comparison:} We streamline the input format to solely consist of pairwise responses. This contrasts with SPCT's capability to handle single or multiple responses. We find that a dedicated pairwise setup simplifies the training process and yields more accurate comparative assessments by eliminating potential distractions from additional responses.
\end{itemize}

The development and training of our Pairwise Writing GenRM, based on the \texttt{Qwen3-32B-Base} model, follows a structured four-step pipeline:
\begin{enumerate}
    \item \textbf{Data Filtering:} Selection of higher-quality data from raw preference datasets.
    \item \textbf{SPCT Prompt Engineering:} Design of tailored SPCT prompts specifically for writing tasks.
    \item \textbf{Cold-Start Fine-tuning:} Collection of a small set of cold-start data using a rejection sampling strategy to initially fine-tune the \texttt{Qwen3-32B-Base} model.
    \item \textbf{RL-based Refinement:} Further enhancement of the GenRM's performance using rule-based reinforcement learning.
\end{enumerate}

\subsubsection{Data Filtering}
\label{data_filter}

Our GenRM training data originates from an in-house dataset of approximately 200K pairwise preferences,
including 30K writing-related pairs.
These writing pairs are re-scored using a scalar RM trained on the full preference set.
We then filter for higher-quality pairs, characterized by a higher chosen response reward and a larger reward gap,
resulting in approximately 10K original writing pairs.
For GenRM evaluation, we use an in-house test set of 1K writing preferences.
The final preference dataset can be described as:
\begin{equation}
\label{common_rule_base_reward}
\mathcal{D} = \{(x^{(i)}, y_c^{(i)}, y_r^{(i)})\}_{i=1}^N
\end{equation}
where $x$ is the query, $y_c$ and $y_r$ are the chosen and rejected responses.

\subsubsection{Writing Principles and Critique}
Our GenRM template, similar to SPCT,
utilizes pre-defined general writing principles and model-generated specific principles
(based on the input query and response pair) to form a detailed critique \citep{liu2025inference}.
The critique concludes with \texttt{\textbackslash boxed\{score\_1, score\_2\}} to denote the comparative quality of the two responses.

The GenRM follows a sequential process: first generating specific principles, then the critique, and finally extracting the paired scores:
\begin{equation}
\{p_i\}_{i=1}^m \sim p_{\theta}\left(x, y_c, y_r\right), \quad \mathcal{C} \sim r_{\theta}\left(x, y_c, y_r, \{p_i\}_{i=1}^m\right), \quad \{S_c, S_r\} = (S_1, S_2) = f_{\text{extract}}(\mathcal{C}),
\end{equation}
where $p_\theta$ denotes the principle generation function (sharing parameters $\theta$ with the reward generation function $r_\theta$), which outputs specific principles $\{p_i\}$.
$\mathcal{C}$ represents the detailed critique generated based on these principles.
$S_c$ and $S_r$ are the individual quality scores for the chosen ($y_c$) and rejected ($y_r$) responses, respectively,
corresponding to $(S_1, S_2)$ based on their presentation order in the template.
$f_{\text{extract}}$ is the function to extract these paired scores from the critique $\mathcal{C}$.
Notably, unlike SPCT, our GenRM outputs $S_c$ and $S_r$ as float numbers within the [0, 10] range, rather than integers.

\subsubsection{Cold Start with RFT}
\label{cold_start}

To mitigate instability in the early stages of RL training and to enhance the granularity of specific principles,
we perform a cold-start fine-tuning (RFT) of the initial RL actor.
This involves curating a small, high-quality dataset as follows:
First, we sample 1k writing preference pairs from the 10k higher-quality pairs in \ref{data_filter}
and swap the positions of chosen and rejected responses to construct 2k SPCT prompts, with 50\% $(y_c, y_r)$ and 50\% $(y_r, y_c)$.
Then we sample initial reasoning traces (specific principles and critiques) for these SPCT prompts using \texttt{Claude-3.5-Sonnet}.
Note that Claude's predictions exhibit a disproportional tendency (around 60\%) to assign higher scores to the former response,
suggesting a potential position bias problem.

Next, we employ a rejection process to refine the dataset.
Specifically, for each SPCT prompt, we examine Claude's predicted $\{S_c, S_r\}$, the pair scores for the chosen ($y_c$) and rejected ($y_r$) responses respectively.
If this prediction does not align with the known ground-truth preference for that pair (i.e., $S_c < S_r$),
the prompt is discarded.
We then proceed to only keep the original queries for which \textit{both} of their corresponding SPCT prompt versions
(i.e., the $(y_c, y_r)$ version and the swapped $(y_r, y_c)$ version) are both consistent with the ground-truth preference.
As a result, we maintain 500 original writing preference pairs, doubling this set to 1,000 pairs by swapping the chosen and rejected responses.
We fine-tune \texttt{Qwen3-32B-Base} with these 1k cold start data for one epoch.

\subsubsection{Rule-based RL for Pairwise GenRM}
\label{genrm_tricks}
The writing GenRM is further fine-tuned by GRPO \citep{shao2024deepseekmath}, with rule-based outcome rewards.
The predicted pair float scores $\{S_c, S_r\}$ are extracted and composed into two kinds of rewards: an accuracy reward and a format reward.

\paragraph{Accuracy Reward}
Formally, for the $i$-th output $o_i$ in the group of $G$ individual responses $\{o_i\}_{i=1}^G$, the accuracy reward is:
\begin{equation}
R_{\text{acc}} =
\begin{cases}
1,  & \text{if } S_c > S_r, \\
-1, & \text{if } S_c < S_r, \\
0,  & \text{otherwise}, \\
\end{cases}
\end{equation}
the rewards are computed for each response $o_i$, we omit the index $i$ for notational simplicity.

\paragraph{Format Reward}
The format reward is also important, for both parsing process and numerical boundary:

\begin{equation}
R_{\text{format}} =
\begin{cases}
1, & \text{if } 0 \leq S_c \leq 10 \text{ and } 0 \leq S_r \leq 10, \\
-1, & \text{otherwise}.
\end{cases}
\end{equation}

\paragraph{Score Margin}
As it is difficult for the writing GenRM to distinguish pair responses with fine-grained textual or semantic divergences,
we introduce a heuristic weighting term to enhance GenRM's sensitivity to these differences:

\begin{equation}
R_{\text{acc}} =
\begin{cases}
\displaystyle R_{\text{acc}} \times \frac{(S_c - S_r)}{\tau_{\text{margin}}}, & \text{if } 0 < S_c - S_r < \tau_{\text{margin}}, \\
R_{\text{acc}}, & \text{otherwise}, \\
\end{cases}
\end{equation}
where the margin threshold is $\tau_{\text{margin}} = 2$ in our implementation.

\paragraph{Position Bias Weight}

Let $P_{\text{former}}$ denote the probability that the GenRM prefers the first-presented (former) response in a pair.
Ideally, we expect $P_{\text{former}}$ to be 0.5 because the training data is deliberately balanced with respect to presentation order,
as it includes swapped chosen-rejected response pairs.
Furthermore, its variance (estimated from batch statistics) should approach 0 on this balanced training data.
The reinforcement learning process helps converge the empirical estimate of $P_{\text{former}}$
(i.e., the proportion of times the former response is preferred in a batch) to approximately 0.5,
yet reducing the variance of these batch-wise proportions remains a challenge.
To mitigate the variance of these batch-wise preference proportions, we apply a weighting to the advantage $\hat{A}_{i,t}$ that is dynamically calculated based on the preference distribution within the current batch:

\begin{equation}
  \hat{A}_{i,t} =
  \begin{cases}
  \hat{A}_{i,t} \times \displaystyle \frac{\text{\#Latter Preference in Batch}}{\text{Batch Size} / 2} , & \text{if } S_c > S_r \text{ and } y_c \text{ was presented first}, \\
  \\
  \hat{A}_{i,t} \times \displaystyle \frac{\text{\#Former Preference in Batch}}{\text{Batch Size} / 2}, & \text{if } S_c > S_r \text{ and } y_c \text{ was presented second}.
  \end{cases}
\end{equation}

\paragraph{Dynamic Sampling}
We employed dynamic sampling following \cite{yu2025dapo}.
This technique involves over-sampling and then filtering out prompts for which the GenRM's predictive accuracy is 0 or 1.

\subsection{RL with Bootstrapped Relative Policy Optimization}
\begin{figure}[htbp]
  \centering
  \begin{subfigure}[b]{0.48\textwidth}
      \includegraphics[width=\textwidth]{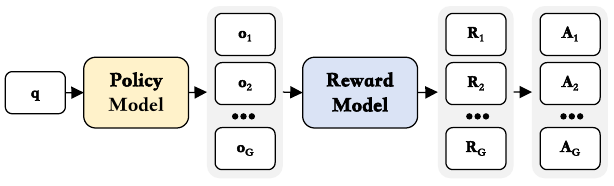}
      \caption{Architecture of GRPO}
      \label{arch_grpo}
  \end{subfigure}
  \hfill 
  \begin{subfigure}[b]{0.44\textwidth}
      \includegraphics[width=\textwidth]{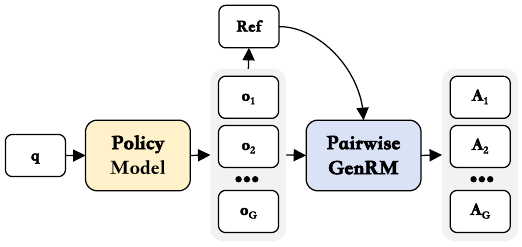}
      \caption{Architecture of BRPO}
      \label{arch_brpo}
  \end{subfigure}
  \caption{Demonstration of GRPO and our BRPO.
  BRPO implements bootstrap by randomly selecting reference response from the current group of policy responses, and achieves zero expectation for the final advantages directly.} 
  \label{fig:arch_brpo}
\end{figure}

\label{writing_model}
In this stage, we introduce Bootstrapped Relative Policy Optimization (BRPO), an algorithm adapted from GRPO \citep{shao2024deepseekmath}.
BRPO is designed to utilize preference rewards obtained from the pairwise GenRM, facilitating robust learning for non-verifiable tasks.
With BRPO, \texttt{Qwen3-32B-Base} is trained as a writing model directly, without supervised fine-tuning (SFT) as a preliminary step,
termed \texttt{Writing Zero}.
Furthermore, we apply BRPO to an in-house SFT model to create \texttt{Writing-R1}.

\subsubsection{Bootstrapped Relative Policy Optimization}
While Group Relative Policy Optimization (GRPO) \citep{shao2024deepseekmath} utilizes group-relative normalization of scalar rewards,
defining such rewards reliably for non-verifiable tasks like creative writing is challenging due to their inherent subjectivity.
Pairwise comparative assessment, however, offers a more robust and scalable evaluation method,
aligning better with human judgments in these subjective domains.
Bootstrapped Relative Policy Optimization (BRPO) is designed to leverage this by enabling dynamic, reference-free pairwise comparison.
It achieves this by utilizing a bootstrapped response—randomly selected from the current group of policy rollouts—as a temporary reference for advantage estimation during RL training, as shown in Figure \ref{arch_brpo}.
This core mechanism eliminates the need for fixed external references, allowing the policy model to continuously improve by learning from its own increasingly sophisticated outputs.
BRPO consists of two main components: a novel \textit{preference advantage} calculation and a \textit{dynamic sampling} strategy.

For a given query, BRPO dynamically establishes a temporary reference by randomly selecting one response,
$o_{\text{ref}}$, from the current group of $G$ policy-generated responses $\{o_i\}_{i=1}^G$.
This $o_{\text{ref}}$ is then used for pairwise comparison against all responses in the group to obtain preference scores.
In other words, BRPO employs a bootstrap approach to iteratively enhance its model capabilities by conducting multiple rounds of voting and calculating preference among its self-generated outputs.
Consequently, our BRPO achieves zero expectation for advantage and requires no additional normalization.

\paragraph{Preference Advantage}
At the core of BRPO's advantage estimation is the direct use of binary pairwise preference signals,
replacing the pointwise scalar normalization found in methods like GRPO (as seen in Equation \ref{grpo_advantage}).
Specifically, for each output $o_i$ within the group of $G$ policy-generated responses $\{o_i\}_{i=1}^G$,
its preference relative to the dynamically selected reference $o_{\text{ref}}$ (as described previously) is determined.
This results in a binary preference score $R_i$, obtained by comparing $o_i$ and $o_{\text{ref}}$ using the pairwise GenRM.
The corresponding scores $S_i$ (for $o_i$) and $S_{\text{ref}}$ (for $o_{\text{ref}}$) are extracted from the GenRM's critique.
The advantage $\hat{A}_{i,t}$ for the $i$-th response at token $t$ is then directly set to this binary preference score:

\begin{equation}
\label{brpo_advantage}
  R_i =
  \begin{cases}
  1, & \text{if } S_i > S_\text{ref} \text{ in all voting conditions} \\
  -1, & \text{otherwise}
  \end{cases}, \quad
   \text{and thus, }
  \hat{A}_{i,t} = R_i.
\end{equation}

\paragraph{Dynamic Sampling}
BRPO also incorporates a dynamic sampling mechanism.
As the policy model $\pi_{\theta}$ evolves during training,
the quality of its generated responses will progress.
A potential issue arises if a chosen $o_{\text{ref}}$ for a particular query happens to be an outlier—either exceptionally strong or weak,
or containing superficial patterns that the GenRM is overly sensitive to.
In such cases, $o_{\text{ref}}$ might be consistently judged as superior (or inferior) to nearly all other responses in the group $\{o_i\}_{i=1}^G$.
This can lead to a batch of preference scores $R_i$ that are predominantly all +1 or all -1,
offering little discriminative signal for policy updates and potentially encouraging the policy to overfit to superficial patterns in $o_{\text{ref}}$.
To mitigate this,
BRPO filters out queries for which the selected $o_{\text{ref}}$ leads to such highly skewed preference distributions within the group.
Formally, for a group of $G$ pairwise preference scores $\{R_i\}_{i=1}^{G}$,
the query $q$ is filtered out from the current training batch if:

\begin{equation}
\frac{ \lvert \sum_{i=1}^{G} R_i \rvert}{G} > \tau_{\text{filter}},
\end{equation}
where $\tau_{\text{filter}}$ is a hyperparameter threshold.

\subsubsection{Writing-Zero and Writing-R1}

Inspired by approaches like DeepSeek-R1-Zero \citep{guo2025deepseek}, which investigate training models from scratch via RL,
we explore a similar paradigm for creative writing with our \textit{Writing-Zero} model.
This involves training \texttt{Qwen3-32B-Base} directly from its pretrained state,
relying exclusively on a pure reinforcement learning process driven by our pairwise writing GenRM
and BRPO algorithm for self-evolution, without task-specific supervised fine-tuning.
Remarkably, Writing-Zero demonstrates the potential to achieve performance comparable to models undergoing traditional SFT and subsequent RL tuning.

To further assess the capabilities of BRPO, we also introduce \textit{Writing-R1}.
For this model, we apply the same BRPO algorithm to fine-tune an in-house SFT model already specialized in thinking and instruction following.
As demonstrated by our empirical results,
Writing-Zero and Writing-R1 not only achieve competitive performance on in-house and open-source writing benchmarks
but also exhibit strong resistance to reward hacking, showcasing the effectiveness of our proposed methodology.

\section{Experiments}

\subsection{Experiment Settings}
\label{exp_settings}

\paragraph{Method Implementation}
The preference data for pairwise writing GenRM is described in Section \ref{data_filter}.
The RL training data for Writing-Zero and Writing-R1 are from in-house unsuperivised queries.
To ensure consistent training and evaluation, we primarily utilize Chinese-based datasets in this work.
Our training framework is based on verl \citep{sheng2024hybridflow}.
We employ vLLM engine \citep{kwon2023efficient} for both GenRM and Writing Model inference.
The learning rate for all training is 1e-6.
For the decoding parameters of both training and evaluation in this paper, \texttt{temperature} is set to 1.0, \texttt{top-p} is set to 1.0, \texttt{top-k} is set to -1.
We set the dynamic sampling threshold $\tau_{\text{filter}}$ to 0.6, which means for a group of 16, if the sum of preference scores is greater than 9, the query will be filtered out.

\paragraph{Eval RM}
To observe the performance changes of the writing model during the BRPO training process,
we have specifically trained a scaler reward model on the in-house writing test dataset, termed Eval RM.
The Eval RM exhibits a high degree of consistency with human evaluation results on the in-house writing test dataset.

\subsection{Benchmarks}
We evaluate our proposed Pairwise Writing GenRM on multiple reward model benchmarks,
demonstrating its effectiveness in enhancing performance during reinforcement learning.

\paragraph{Reward Model Benchmarks}
We compare our Pairwise Writing GenRM with Skywork-Reward \citep{liu2024skywork}, INF-ORM \citep{INF-ORM-Llama3.1-70B},
our in-house Writing Scalar RM (trained with 200K full preference data in \ref{data_filter}),
LLM-as-a-Judge baselines (Claude-3.5-Sonnet), across a range of established reward model benchmarks.
RewardBench \citep{lambert2024rewardbench} is one of the earliest benchmarks employing prompt–chosen–rejected triplets for reward model evaluation,
encompassing diverse tasks in chat, reasoning, and safety, with approximately 3,000 annotated examples.
M-RewardBench \citep{gureja2024m} extends this benchmark to a multilingual setting.
To better assess our model’s discrimination ability in Chinese, we restrict evaluation to its Chinese subset.

Additionally, given that our reward model is primarily designed for Chinese writing scenarios,
we construct two additional domain-specific datasets: Cultural \& Creative Writing and Lifestyle Copywriting.
The former is a general-purpose creative writing dataset that includes tasks such as essay composition,
workplace copywriting, and literary creation, comprising a total of 1,036 samples.
The latter focuses on common everyday writing scenarios—such as chatting with friends,
composing text messages, or posting on social media—and includes 831 samples.

\paragraph{Writing Benchmarks}
We evaluate the writing capability of our models and compare their performance with other baseline models,
including Qwen3-32B-Base, Qwen3-32B-Instruct, DeepSeek-R1, Writing-SFT (our in-house thinking SFT model).
WritingBench \citep{wu2025writingbench} serves as a comprehensive benchmark that covers 6 core domains and 100 subdomains,
encompassing a wide range of writing tasks and styles.

In addition, to better align with our real-world application scenarios,
we curated a diverse set of 211 user queries, named as Writing Testset.
Based on history responses and human annotations, we construct the preference dataset,
from which we trained the Eval RM in \ref{exp_settings}.
Eval RM is used to score the responses generated by different models, providing a task-specific evaluation of writing quality.
We observed a strong positive correlation between the scores produced by this reward model and human judgments,
indicating its effectiveness and reliability in assessing writing performance.

\subsection{Main Results}
\subsubsection{Reward Model Results}

\begin{table}[htp]
  \centering
  \resizebox{\textwidth}{!}{
    \begin{tabular}{lcccc}
      \toprule
      \textbf{Model} & \textbf{Cultural \& Creative Writing} & \textbf{Lifestyle Copywriting} & \textbf{Reward Bench} & \textbf{M-RewardBench} \\
      \midrule
      Skywork-Reward-Gemma-2-27B-v0.2 & 56.4 & 56.3 & 94.3\textsuperscript{*} & 90.8 \\
      INF-ORM-Llama3.1-70B & 56.6 & 54.0 & \textbf{95.1}\textsuperscript{*} & \textbf{91.4} \\
            Writing Scalar RM & \textbf{64.7} & \textbf{62.8} & 92.9 & 89.4 \\
      \midrule
      Claude-3.5-Sonnet & 53.0 & 46.8 & 84.2\textsuperscript{*} & 79.7 \\
            Pairwise Writing GenRM & 57.5 & 54.8 & 87.4 & 86.1 \\ 
      \bottomrule
    \end{tabular}
  }
  \vspace{0.5em}
  \caption{Performance of different types of reward models on benchmark datasets.
  The superscript asterisk (*) indicates results taken directly from the RewardBench Leaderboard.}
  \label{rm_result}
\end{table}

As shown in Table \ref{rm_result}, evaluated on the same prompt sets,
our Pairwise Writing GenRM outperforms Claude-3.5-Sonnet across all four benchmarks.
Surprisingly, despite being trained exclusively on Chinese writing data
(without any English or general-domain examples), it achieved 87.4\% accuracy on RewardBench
and 86.1\% on M-RewardBench, showcasing strong generalization performance beyond its primary training scope.
Furthermore, on the Cultural \& Creative Writing dataset, our Pairwise Writing GenRM
achieved 1.1\% and 0.9\% higher accuracy than Skywork-Reward-Gemma-2-27B-v0.2
and INF-ORM-Llama3.1-70B, respectively.
While its performance is lower than our in-house Writing Scalar RM, 
this is anticipated given the reported version of Pairwise GenRM was trained on significantly less data.

\begin{figure}[htbp]
  \centering
  \begin{subfigure}[b]{0.48\textwidth}
      \includegraphics[width=\textwidth]{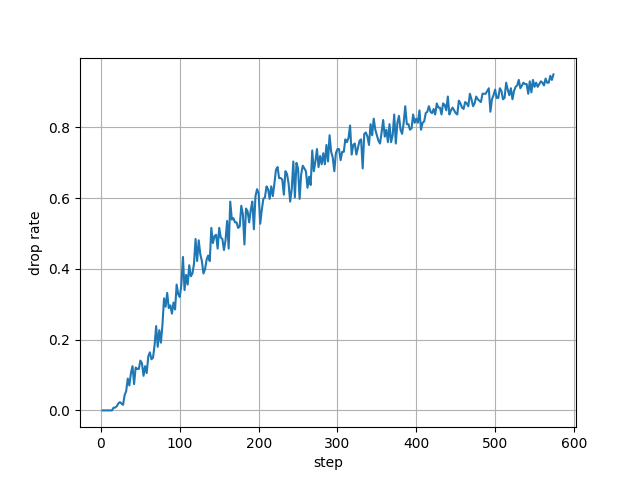}
      \caption{Drop rate of GenRM on train data}
      \label{genrm_drop_rate}
  \end{subfigure}
  \hfill 
  \begin{subfigure}[b]{0.48\textwidth}
      \includegraphics[width=\textwidth]{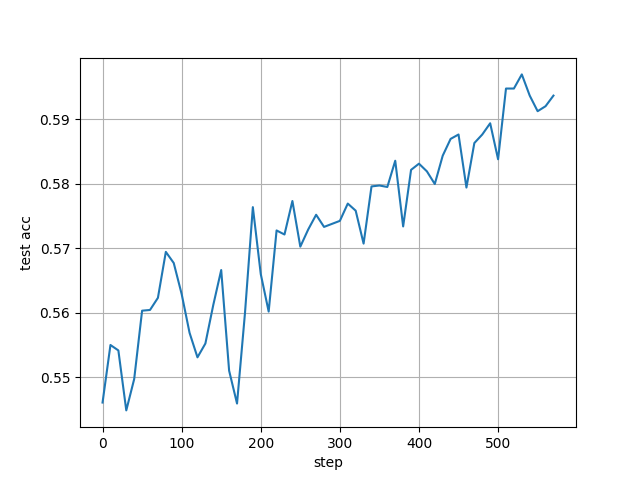}
      \caption{Accuracy of GenRM on test data}
      \label{genrm_test_acc}
  \end{subfigure}
  \caption{GenRM training dynamics, showing test accuracy and training data drop rate.}
  \label{fig:genrm_training}
\end{figure}

While the Pairwise Writing GenRM demonstrated promising benchmark results,
its training process encountered a notable challenge.
Specifically, during GenRM training, dynamic sampling caused a significant computational bottleneck.
After certain iterations, the dropout rate progressively increased, exceeding 95\% in the later stages
(as shown in Figure \ref{fig:genrm_training}a).
This extremely high dropout rate caused severe computational inefficiency, as
a large number of rollouts are required to assemble enough training batch,
making further training prohibitively slow.
Consequently, we had to halt the GenRM training process prematurely.
As a result, the model's accuracy on the test set did not show clear signs of convergence
(Figure \ref{fig:genrm_training}b), implying that the GenRM employed in subsequent RL stages
was likely sub-optimal and that further refining its training could enhance overall RL performance.
We plan to address this issue in future work.

\subsubsection{Writing Model Results on Writing Benchmarks}

\begin{table}[htp]
  \centering
  \small
  \begin{tabularx}{\linewidth}{lcccc}
  \toprule
  \textbf{Model} & \textbf{WritingBench} & \textbf{Writing Testset}  \\
  \midrule
  Qwen3-32B-Base      & 6.89 & 1.23  \\
  Qwen3-32B-Base-ScalarRM-GRPO      & \sout{8.87} & 2.83 \\
  Qwen3-32B-Base-GenRM-BRPO / \textbf{Writing-Zero}      & 8.29 & 3.84 \\ 
  Qwen3-32B-Base-GenRM-BRPO / \textbf{Writing-Zero} (voting@2)      & 8.35 & 3.86 \\ 

  \midrule
  Qwen3-32B-Instruct      & 8.64 & 3.32 \\  
  Deepseek-R1      & 8.55\textsuperscript{*} & 3.20\\
  Writing-SFT      & 8.56 & 3.31 \\           
  Writing-SFT-ScalarRM-GRPO      & \textbf{8.77} & 3.69 \\
  Writing-SFT-GenRM-BRPO / \textbf{Writing-R1}      & 8.68 & \textbf{3.93}  \\
  \bottomrule
  \end{tabularx}
  \vspace{0.5em}
  \normalsize
  \caption{Performance of different models on writing-related evaluation datasets.
    The * indicates results taken directly from the original benchmark.
  Strikethrough results (e.g., \sout{8.87}) marks scores potentially inflated by reward hacking.
  }
  \label{writing_model_res}
\end{table}

As shown in Table \ref{writing_model_res},
RL training substantially improves the writing capabilities of both the \texttt{Qwen3-32B-Base} and our in-house \texttt{Writing-SFT} model.
Our core approach involves training these base models with Pairwise Writing GenRM using the BRPO algorithm,
resulting in \textbf{Writing-Zero} (\texttt{Qwen3-32B-Base-GenRM-BRPO}) and \textbf{Writing-R1} (\texttt{Writing-SFT-GenRM-BRPO}).
For comparison, we also implemented baselines where these base models were trained with a Scalar Reward Model (ScalarRM) using the GRPO algorithm,
denoted as \texttt{Qwen3-32B-Base-ScalarRM-GRPO} and \texttt{Writing-SFT-ScalarRM-GRPO}.

Notably, when guided by our Pairwise GenRM with BRPO:
\begin{itemize}
    \item \textbf{Writing-Zero} improves from the base Qwen3-32B-Base scores of 6.89 to 8.29 on WritingBench and from 1.23 to 3.84 on the Writing Testset.
    \item \textbf{Writing-R1} (based on Writing-SFT) shows gains to 8.68 on WritingBench (from Writing-SFT's 8.56) and to 3.93 on the Writing Testset (from Writing-SFT's 3.31).
\end{itemize}
Despite the Pairwise GenRM's lower raw performance on some reward model benchmarks (Table \ref{rm_result}) compared to the Scalar RM,
it demonstrates greater efficacy when integrated into the RL process with BRPO.
As depicted in Figure \ref{fig:writing_zero}, Writing-Zero
not only achieves higher and more stable reward scores during training compared to \texttt{Qwen3-32B-Base-ScalarRM-GRPO},
but also translates to better final performance on the Writing Testset, surpassing it by 1.01 points.
Similarly, Writing-R1 outperforms \texttt{Writing-SFT-ScalarRM-GRPO} by 0.24 points on the same test set.

A crucial observation pertains to reward hacking: during the training of \texttt{Qwen3-32B-Base-ScalarRM-GRPO},
we noted early instances where its responses devolved into gibberish and became largely unreadable.
This poor quality was reflected in its low Eval RM scores on our Writing Testset.
However, this same model achieved an anomalously high score on WritingBench,
suggesting WritingBench's susceptibility to such hacking, a vulnerability our Eval RM resists.

To further validate these automated metrics, we conducted human evaluations.
On a test set of 166 instances, Writing-R1 demonstrated superior performance
when compared against both its SFT base (G:S:B ratio of 47:106:13 in favor of Writing-R1 vs. Writing-SFT)
and the scalar-reward trained counterpart (G:S:B ratio of 28:120:18 in favor of Writing-R1 vs. \texttt{Writing-SFT-ScalarRM-GRPO}).

\section{Analysis}
\subsection{Reward Hacking in Writing}

\begin{table}[htp]
  \centering
  \resizebox{\textwidth}{!}{
    \begin{tabular}{lcccc}
      \toprule
      \textbf{Model} & \textbf{Mean Response Len} & \textbf{Mean Explanation Len} \\
      \midrule
      Qwen3-32B-Base-ScalarRM-GRPO & 1872 & 417 \\
      Qwen3-32B-Base-GenRM-BRPO / \textbf{Writing-Zero} & 1292 & 58 \\
      Writing-SFT                   & 1251 & 125 \\
      \bottomrule
    \end{tabular}
  }
  \vspace{0.5em}
  \caption{Mean length of response and redundant explanation.}
    \label{reward_hack_table}
\end{table}

Creative writing tasks are susceptible to two significant reward hacking issues:
1) \textit{Length bias}: Reward models often assign higher scores to longer responses, leading RL-trained models to become excessively verbose.
2) \textit{Redundant explanations}: Reward models may favor responses with lengthy, often unnecessary, self-justifications or praise appended to the actual content, causing RL-trained models to generate such superfluous text.

Reward hacking results in superficial improvements in reward metrics during training, as the actor model learns to exploit loopholes in the reward model rather than acquiring genuine writing proficiency.
To quantify these effects,
we analyzed the performance of three key models on these hacking issues using an internal creative writing test set.
As shown in Table \ref{reward_hack_table},
our \textbf{Writing-Zero} model demonstrates substantially less redundant information
and shorter response lengths compared to \texttt{Qwen3-32B-Base-ScalarRM-GRPO}, highlighting its improved resistance to these hacking phenomena.

\subsection{GenRM's Position Bias}

\begin{figure}[htbp]
  \centering
  \begin{subfigure}[b]{0.48\textwidth}
      \includegraphics[width=\textwidth]{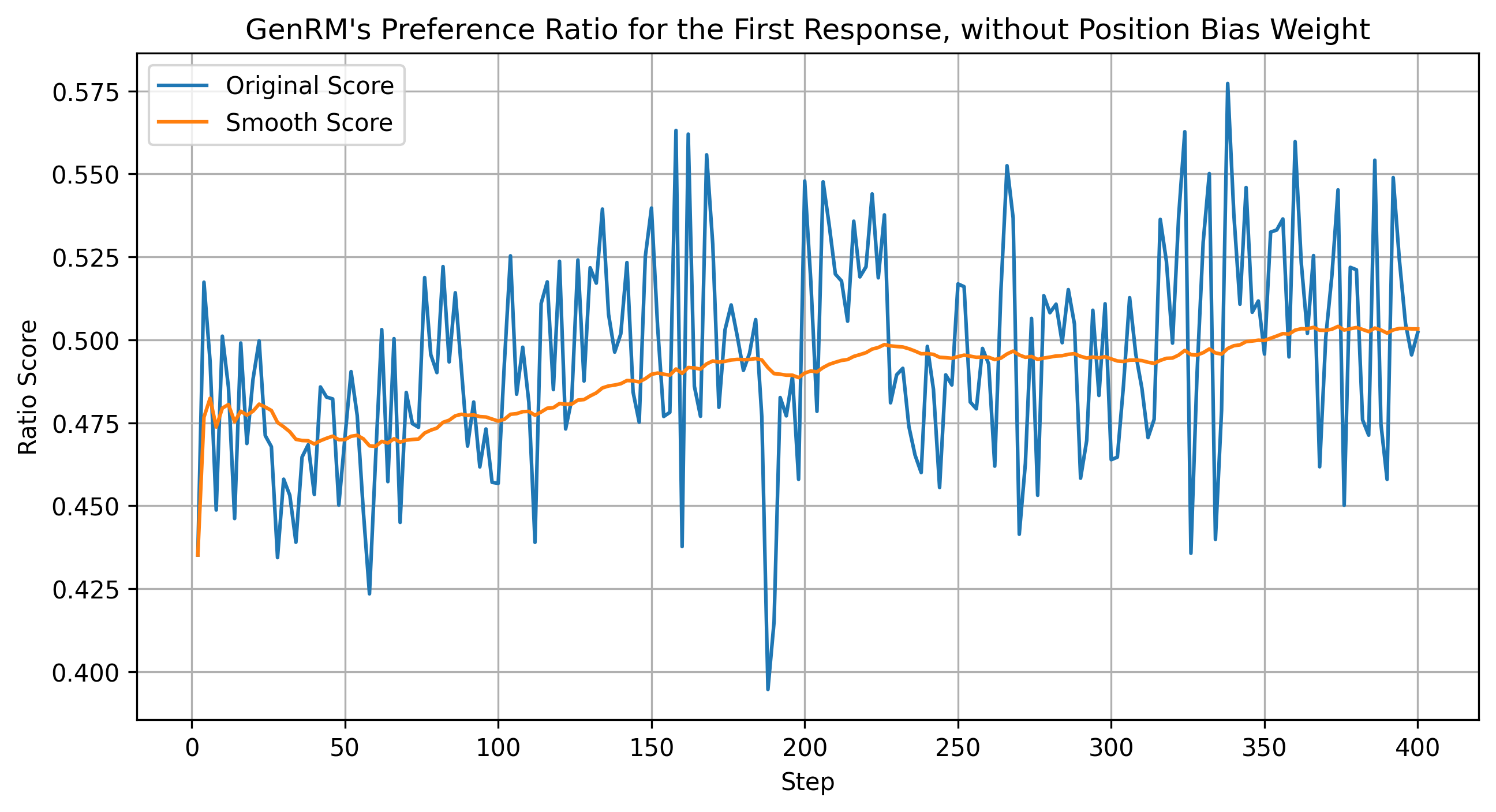}
      \caption{Preference Ratio without Position Bias Weight}
      \label{ratio_without_position_bias}
  \end{subfigure}
  \hfill 
  \begin{subfigure}[b]{0.48\textwidth}
      \includegraphics[width=\textwidth]{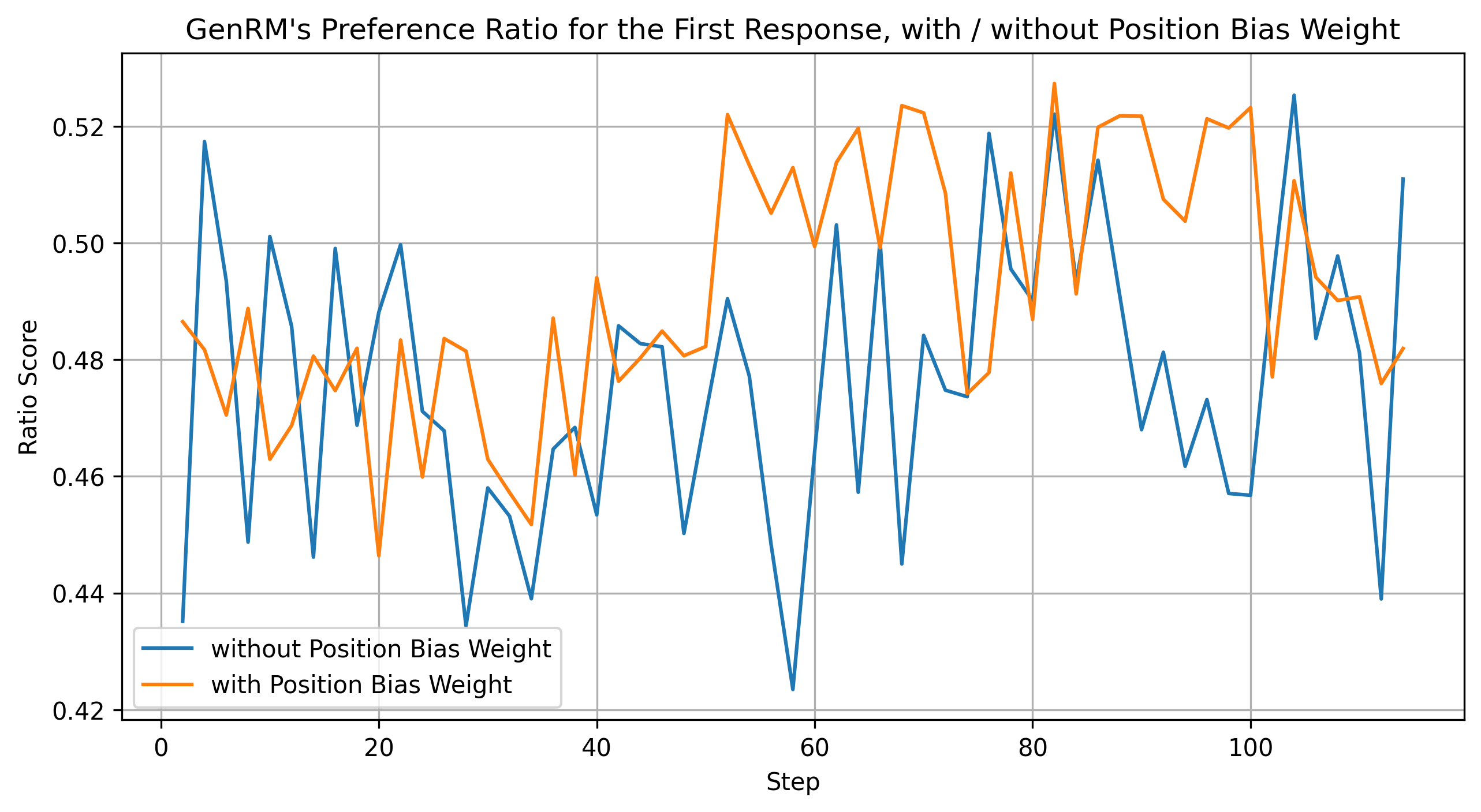}
      \caption{Preference Ratio with Position Bias Weight}
      \label{ratio_with_without_position_bias}
  \end{subfigure}
  \caption{Convergence of the GenRM's preference ratio during RL training.} 
  \label{fig:position_bias_convergence} 
\end{figure}

After the cold-start fine-tuning (RFT) detailed in Section \ref{cold_start},
our initial fine-tuned GenRM exhibited a significant position bias.
Unlike Claude-3.5-Sonnet, which showed a ~60\% tendency to favor the first response,
our model initially preferred to assign higher scores to the latter response in a pair.
Fortunately, this inherent bias was largely calibrated during the subsequent RL phase.
As illustrated in Figure \ref{ratio_without_position_bias},
the empirical probability of the GenRM preferring the former response gradually converged towards the ideal 50\% during RL training.

While the mean of this preference probability (as shown in Figure \ref{ratio_without_position_bias}) converged towards 0.5,
its variance across training batches did not show a similar convergent tendency without intervention.
As demonstrated in Figure \ref{ratio_with_without_position_bias},
the position bias advantage weighting mechanism visibly reduced the variance of the preference ratio.

\begin{figure}[h]
  \centering
  \includegraphics[width=0.5\linewidth]{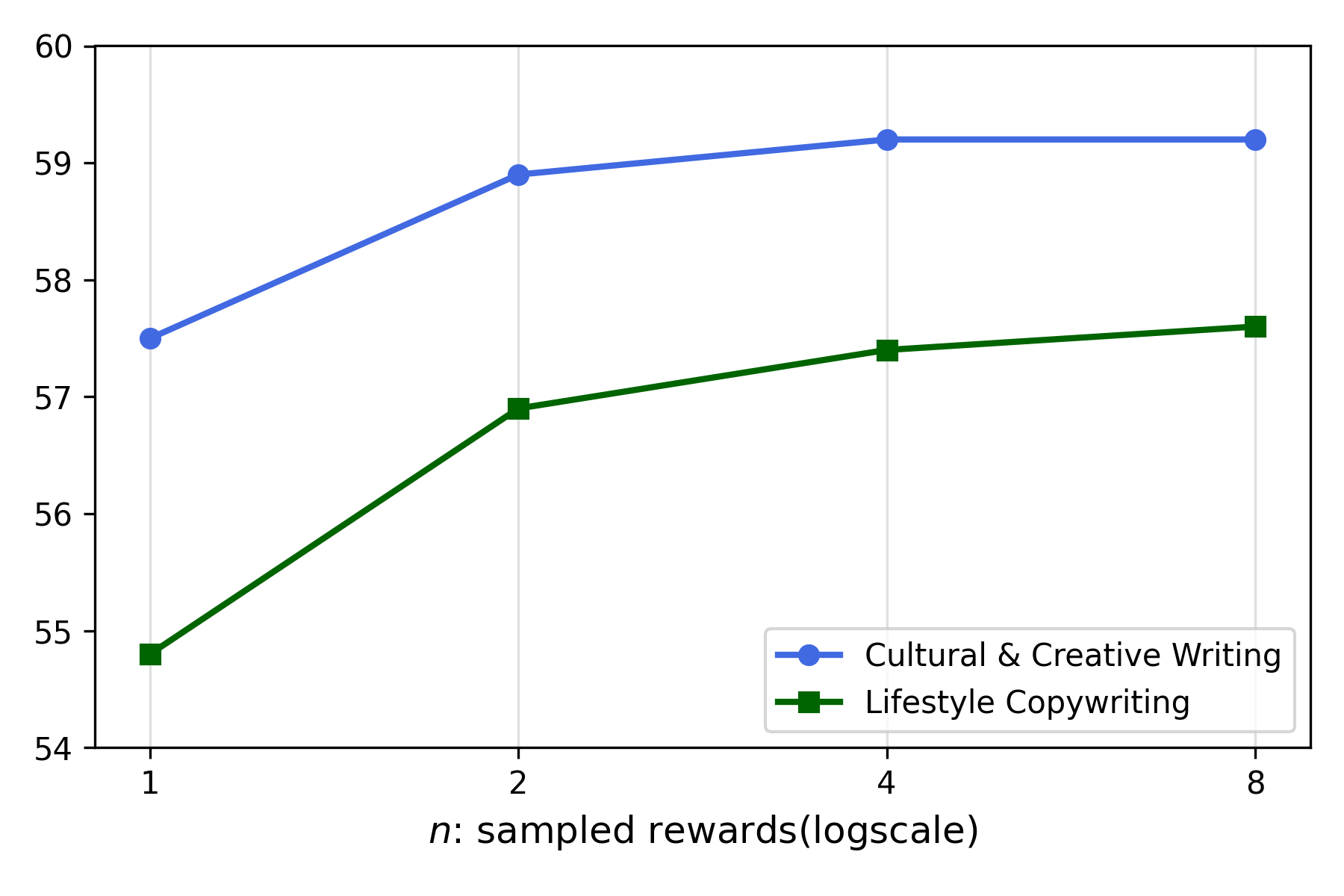}
  \caption{Majority voting accuracy (voting@$n$) across $n = 1, 2, 4, 8$ on internal datasets, as evaluated by the pairwise GenRM.}
  \label{genrm_vote}
\end{figure}

\subsection{Test-time Scaling Benefits}
A notable advantage of GenRMs over scalar reward models is their capability for test-time scaling
through majority voting over multiple generations.
This can enhance both the direct accuracy of the GenRM itself and, consequently, the performance of the policy model trained with it.

As illustrated in Figure \ref{genrm_vote},
increasing the number of voting samples ($n$) directly improves the accuracy of our Pairwise Writing GenRM on our internal evaluation datasets.
For instance, with $n=8$ votes, accuracy on the Cultural \& Creative Writing dataset increases from 57.5\% (for $n=1$) to 59.2\%, and on the Lifestyle Copywriting dataset from 54.8\% to 57.6\%. This demonstrates the GenRM's ability to refine its judgments by aggregating multiple perspectives.
The benefits of test-time scaling extend to the policy models trained with the GenRM.
As shown in Table \ref{writing_model_res}, when \textbf{Writing-Zero} utilizes its Pairwise GenRM with two voting rounds (voting@2),
it achieves further improvements, which indicates that a more accurate reward signal, refined by test-time scaling,
can lead to better policy optimization and ultimately stronger performance on downstream writing tasks.
Further enhancing the GenRM's performance could unlock greater test-time scaling benefits, for example, by enabling strategies for multiple voting rounds (n > 1) that selectively reward consistently superior responses while neutralizing or penalizing inconsistently judged ones, thereby fostering more robust policy learning.

We believe that with further improvements to the GenRM's accuracy,
the potential of test-time scaling with multiple voting rounds could be more deeply explored.
For instance, an open question for future experimentation is whether responses with inconsistent judgments across multiple votes should be neutrally rewarded (e.g., 0) or negatively penalized (e.g., -1 in Equation \ref{brpo_advantage}).

\subsection{Intuition behind reference selection of BRPO}
The primary challenge in non-verifiable tasks like creative writing is the absence of definitive ground-truth answers,
making quality assessment inherently relative and reliant on comparison.
We initially considered using the best-performing or relatively better response from a previous group rollout as reference for the next training iteration.
However, this approach proved problematic.
As the policy model's capabilities significantly improve during training,
such a static or outdated reference quickly becomes suboptimal,
introducing an "offline" data issue that can impede further learning by comparing against a weaker baseline.
Furthermore, our pairwise GenRM is designed for relative comparisons rather than assigning absolute scores.
Attempting to derive a consistent listwise ranking from purely pairwise preferences would necessitate an additional redundant pointwise reward model.
Consequently, BRPO adopts a dynamic, bootstrapped reference selection strategy,
where a randomly chosen response from the current policy rollouts serves as a temporary,
relevant reference for pairwise comparisons within the same group,
fostering continuous self-improvement without external anchors.

\section{Related Work}
\subsection{Reward Model}
Early approaches for reward modeling primarily focused on discriminative reward models,
which were trained to directly assess the quality of individual responses \citep{liu2024skywork,cai2024internlm2}.
These pointwise assessment methods laid the groundwork for preference learning,
but frequently faced challenges with generalization across diverse contexts and over-optimization \citep{yang2024regularizing,sharma2023towards,mckee2024honesty}.
Another line of work gives pairwise scores to evaluate whether responses match the reference answer,
or compare the quality of two responses \citep{xu2025unified,jiang2023llm},
showing that by learning to predict which of two responses is preferred,
generally leads to improved stability and better capture the relative nuances of human preferences compared to direct scalar assessments.
Generative Reward Models (GenRMs) introduce a new paradigm \citep{li2023generative,vu2024foundational,ye2024beyond,zhang2024generative,cao2024compassjudger},
which allows the reward model to leverage chain-of-thought and even use test-time scaling to make more reliable evaluation.
\cite{liu2025inference} stand out as the most relevant works to our work.
They introduced a novel learning method enabling GenRMs to generate adaptive principles and accurate critiques
to give a more reliable comparison between two responses through a two-phase training process: distillation then RLVR.
\cite{chen2025rm,whitehouse2025j1,chen2025judgelrm} share a similar idea of using designed evaluation criteria and RLVR to improve the quality of GenRM.
\cite{zhang2025r1,wang2025unified} also expand GenRM and RLVR to multimodal domains.
In this work, we explore the application of pairwise GenRM to non-verifiable writing tasks.

\subsection{Reinforcement Learning}
Reinforcement Learning from Human Feedback (RLHF) has become a crucial component
for aligning Large Language Models (LLMs) with human preferences and desired behaviors \citep{ouyang2022training}.
Recently, Reinforcement Learning with Verifiable Rewards (RLVR)
has proven an effective paradigm for improving the reasoning capabilities of LLMs in domain such as math and coding \citep{jaech2024openai,guo2025deepseek,team2025kimi},
inspiring subsequent research to actively explore various potentially more advanced reasoning RL algorithms \citep{yu2025dapo,xiong2025minimalist,li2025adaptive}.
\cite{su2025expanding} and \cite{xu2025unified} propose to expand RLVR to problems with unstructured reference,
which often rely on a pre-defined fixed reference or a best-of-n responses sampled from an initial or static model,
to serve as a ground-truth reference.
In this work, our proposed Bootstrapped Relative Policy Optimization (BRPO) algorithm is a step further to explore the application of RLVR to non-verifiable tasks.

\section{Conclusion, Limitations, and Future Work}
In this work, we propose a unified RLVR-based training paradigm to bridge the gap between non-verifiable tasks and verifiable rewards.
By introducing a writing-principle-based pairwise Generative Reward Model and the Bootstrapped Relative Policy Optimization algorithm,
our approach enables large language models to achieve stable and scalable reinforcement learning, even in highly subjective domains such as creative writing.
Experimental results demonstrate that our method not only enhances robustness against reward hacking but also yields competitive performance across different writing benchmarks.
This work lays the foundation for a comprehensive and consistent RL framework that is applicable across the full spectrum of language tasks.
However, due to limited computational resources, we were unable to systematically analyze the trends of test-time scaling of our pairwise writing GenRM, nor could we fully investigate the impact of various parameters and settings within both the GenRM and RL stages.
We also leave the integration of all reward modeling paradigms in a unified framework—along with the extension to multi-modal scenarios—as directions for future research.

\bibliographystyle{unsrtnat}
\bibliography{citations}

\appendix

\newpage


\section*{Supplementary material}

# A Case Study

> **Query 1**
>
> 请为陪伴 15 年的猫咪设计告别卡片文字，要体现我和猫咪的深厚感情

> **Response: Writing-SFT-Scalar-RM-RLHF**
>
> <think>
> 用户让我为陪伴 15 年的猫咪设计告别卡片的文字，要体现他们之间的深厚感情。首先，我需要理解用户的需求。15 年的陪伴，这相当于猫咪几乎贯穿了用户的大半个人生，所以感情一定非常深厚。用户可能希望卡片文字能够表达感激、怀念以及猫咪在他们生活中的重要性。
>
> 接下来，我得考虑用什么样的语气和风格。告别卡片通常需要温暖、感人，同时带有一些回忆的元素。可能需要用一些比喻，比如家人、朋友、伙伴，或者用时间相关的词汇，比如"15 年"。另外，猫咪的特点有没有提到？比如名字、习惯、共同经历，但用户没有提供这些细节，所以可能需要保持一定的通用性，但又要让用户觉得个性化。
>
> 然后，结构方面，可能需要分几个部分：开头表达离别的不舍，中间回忆共同度过的时光，最后表达感谢和祝福。可能需要用一些诗意的句子，或者排比结构，让文字更有感染力。
>
> 还要注意避免过于悲伤，虽然告别难免伤感，但也要有温馨的回忆，让整体感觉既有哀思又不失美好。可能需要加入一些具体的场景，比如晒太阳、蹭腿、夜晚的陪伴，这些细节能让文字更生动，引起共鸣。
>
> 另外，用户可能需要中英文版本，或者不同的风格选择，比如经典、温馨、诗意等，所以可能需要提供几种不同的模板，让用户可以根据自己的喜好选择。但用户当前的要求可能更偏向中文，所以需要确保文字符合中文的表达习惯，用词恰当，避免生硬。
>
> 还要考虑卡片的格式，比如是否需要分标题、正文、结尾，或者是否需要加入一些符号，比如爱心、爪印，这些可能在设计卡片时会用到，但文字部分需要简洁有力。可能需要建议用户搭配一些图案，但主要任务是文字内容。
>
> 最后，检查是否有遗漏的情感点，比如猫咪带来的安慰、陪伴的意义、对未来的祝福，或者对猫咪在另一个世界的美好祝愿。这些元素都能增强文字的深度和情感。
>
> 总结下来，我需要写出几个版本的文字，涵盖不同的风格，同时确保每一版都突出 15 年的漫长陪伴和深厚的感情，使用具体的回忆细节，温暖而感人的语气，结构清晰，让读者感受到用户与猫咪之间不可磨灭的纽带。
> </think>
>
> 以下是为陪伴 15 年的猫咪设计的告别卡片文字，分为不同风格，体现深厚感情和珍贵回忆：
>
> ---
>
> **1. 温馨叙事版 **
> 亲爱的 [猫咪名字]：
> 十五年，你从毛茸茸的小团子变成我生命里不可或缺的家人。那些清晨的蹭脸问候、午后的阳光共眠、深夜的温柔陪伴……你总能用咕噜声治愈我的疲惫。你教会我耐心与爱，陪我度过欢笑与泪水。如今，虽然你要去喵星的彩虹桥，但我们的回忆会永远温暖我的心。谢谢你，我的挚友，愿你下一程旅途充满小鱼干与温柔的风。
> 永远爱你的 [你的名字]
>
> **2. 诗意抒情版 **
> 致我的毛孩子 [猫咪名字]：
> 十五载春秋，你是时光的精灵，
> 用软糯的喵语编织爱的经纬。
> 窗台的阳光有你的形状，
> 深夜的呼吸有你的频率。
> 你曾是襁褓中的绒毛星，
> 后来成了岁月里的静默诗。
> 如今星辰召唤你归去，
> 但我们的羁绊早化作银河的絮语。
> 喵星没有病痛，只有无尽的鱼宴与软垫，
> 我会在月光下，继续读你留下的温柔笔记。
>
> **3. 感恩回忆版 **
> 亲爱的 [猫咪名字]：
> 十五年，足够让一棵树苗长成大树，
> 也足够让一份陪伴沉淀成永恒。
> 谢谢你陪我走过青春到中年的晨昏，
> 用无言的守候填补生命的缝隙。
> 你的顽皮、你的温柔、你独特的呼噜歌，
> 都成了记忆里最珍贵的宝藏。
> 虽然眼泪会模糊离别，但微笑属于我们共度的每一天。
> 喵星的路很远，但爱会一直照亮方向。
>
> **4. 经典简约版 **
> 致最爱的 [猫咪名字]：
> 十五年，一生相伴。
> 谢谢你，我的小猫，我的家人。
> 愿喵星再无病痛，只有安眠。
>
> ** 设计思路解析：**
> - ** 时间维度 **：强调"15 年"强化漫长陪伴带来的深厚情感。
> - ** 回忆具象化 **：用"蹭脸问候、阳光共眠、咕噜声"等细节唤醒共同记忆。
> - ** 情感升华 **：将猫咪定位为"家人、挚友、时光的精灵"，超越宠物范畴。
> - ** 温柔祝福 **：对喵星的美好想象，缓解离别的悲伤，突出爱与怀念。
>
> 可根据你和猫咪的独特经历调整细节，让文字更贴合专属回忆。希望这份告别充满温暖与治愈。

15

> **Response: Writing-SFT-GenRM-RLHF**
>
> <think>
> 用户让我为陪伴15年的猫咪设计告别卡片的文字，要体现他们之间的深厚感情。首先，我需要理解用户的需求。15年的陪伴，这相当于猫咪几乎贯穿了用户的大半个人生，所以感情一定非常深厚。用户可能希望卡片文字能够表达感激、怀念以及猫咪在他们生活中的重要性。
>
> 接下来，我得考虑用什么样的语气和风格。告别卡片通常需要温暖、感人，同时带有一些回忆的元素。可能需要用一些比喻，比如家人、朋友，或者用时间相关的词汇，比如"15年"来强调时间的长度。
>
> 然后，我要考虑结构。通常卡片会有开头、中间的回忆部分、表达感谢，以及结尾的告别和祝福。可能需要分几个段落，每段集中一个主题，比如回忆共同经历，感谢陪伴，以及最后的道别。
>
> 还要注意用词的选择，要避免过于悲伤，而是更多地强调美好的回忆和猫咪带来的快乐。比如用"温暖"、"伙伴"、"朋友"、"家人"这些词，能传达出积极的情感。同时，可能需要加入一些具体的回忆，比如午后的阳光、温柔的呼噜声，这些细节能让文字更生动，更有个性化。
>
> 另外，用户可能希望文字简洁但有力，不需要太长，但要足够感人。可能需要使用一些排比句或者重复的结构来加强情感，比如"谢谢你陪我过……谢谢你在我……谢谢你用……"。这样的结构能突出感谢的不同方面。
>
> 还要考虑是否需要加入一些诗意或者象征性的元素，比如用"小天使"、"星辰"来比喻猫咪，或者用"彩虹桥"这样的常见宠物告别意象，让卡片更有意境。
>
> 最后，检查是否符合用户的所有要求：15年的时光、深厚的感情、温暖感人的语气、具体的回忆点、结构清晰。可能需要调整段落顺序，确保情感流动自然，从回忆到感谢再到告别，层层递进。
>
> 可能还需要考虑用户的文化背景，比如是否喜欢某些比喻，或者是否有特定的宗教信仰，但用户没有提到，所以保持通用。同时，避免使用可能引起不适的词汇，比如"死亡"可能用"离开"、"去了喵星球"等更温和的说法代替。
>
> 总结下来，结构大致分为：开头点明时间，中间回忆具体场景和感谢，结尾告别和祝福。用词温暖，充满感情，加入具体细节和比喻，使文字更具感染力。
> </think>
>
> 以下是为陪伴15年的猫咪设计的告别卡片文字，力求温暖、真挚，体现深厚感情：
>
> ---
>
> **致我的挚友小天使：**
>
> 十五年，足够让时光在指尖流淌成河，
> 而你，始终是我世界里最温柔的常数。
>
> 记得你蜷在书页间的午后，
> 记得你蹭过脚踝的清晨，
> 记得你总在深夜用呼噜声，
> 为我缝补那些疲惫的梦境。
>
> 你不仅是猫咪，更是我的家人，
> 用毛茸茸的耐心教会我：
> 爱，是无声的陪伴，是永恒的注视，
> 是即使不说，也懂的默契。
>
> 如今，你轻轻走向彩虹桥的另一端，
> 请带着我们共同的记忆前行——
> 那些晒过阳光的窗台、偷吃的零食、还有数不清的拥抱。
>
> 谢谢你陪我度过青春的莽撞与成年的孤独，
> 谢谢你用一生温柔，治愈我的每一个角落。
>
> 小天使，我会永远记得你，
> 尾巴摇晃的节奏、眼睛里的星辰，
> 以及你存在本身，就是最美好的奇迹。
>
> 愿喵星球有吃不完的小鱼干，
> 有永远晒不烫的懒人沙发，
> 最重要的是，永远记得回家的路。
>
> ——永远爱你的铲屎官
>
> **设计思路：**
>
> 1. **时间维度**：以"十五年"开篇，强调漫长岁月中猫咪的恒定陪伴。
> 2. **情感具象化**：通过"书页间""脚踝""深夜"等细节，唤起共同回忆。
> 3. **身份升华**：将猫咪定义为"家人""挚友"，突出超越宠物的情感纽带。
> 4. **温柔告别**：用"彩虹桥""喵星球"等意象替代离别，减少沉重感，侧重美好祝福。
> 5. **永恒记忆**：结尾强调猫咪的存在本身就是奇迹，将感情升华为珍藏心底的永恒印记。
>
> ---
>
> 可根据猫咪的具体性格或共同经历，加入更个性化的细节（如昵称、特殊习惯等），让文字更具专属感。

16

### Query 2

作为《罗马假日》影迷，在西班牙台阶拍摄打卡照时需要配文，要求含蓄致敬经典但避免直接引用台词

### Response: Writing-SFT-Scalar-RM-RLHF

<think>
用户是《罗马假日》的影迷，想在西班牙台阶拍打卡照配文，要求含蓄致敬经典但避免直接引用台词。首先，我需要理解用户的需求。他们希望配文既有对电影的致敬，又不显得太直白，避免使用电影中的原句。

接下来，我得回忆一下《罗马假日》中的经典元素。西班牙台阶是电影中的重要场景，安妮公主在这里吃冰淇淋，和乔相遇。所以配文需要提到这些元素，比如冰淇淋、罗马的浪漫、偶遇、自由等主题。

然后，用户可能希望配文简洁且有诗意，适合社交媒体发布。可能需要使用一些比喻，比如将台阶比作时光隧道，或者用冰淇淋作为象征，代表短暂的甜蜜和自由。

还要考虑避免直接引用台词，比如"I have to leave you now"或者"Roman Holiday"，所以需要用隐喻的方式表达。比如"罗马的日光明媚，冰淇淋融化在指尖，而故事，永远留在台阶上"这样的句子，既点出了冰淇淋和台阶，又暗示了电影中的浪漫瞬间。

另外，用户可能希望配文适合不同的拍照风格，无论是复古风还是现代风，所以文字需要有一定的通用性。可能需要加入一些旅行或自由的元素，比如"在西班牙台阶，偶遇一场罗马假日"或者"台阶上的每一步，都是时光的礼物"。

还要注意语言的含蓄性，让读者需要稍微联想才能想到电影，而不是直接说出来。比如"某个夏日午后，台阶上的冰淇淋与浪漫永不落幕"这样的句子，既点出了时间和场景，又带出了浪漫的氛围。

最后，可能需要提供几个不同的版本，让用户可以根据自己的照片风格和心情选择。比如文艺风、轻松活泼、复古感、隐喻风格等，满足不同的表达需求。
</think>

**1. 文艺风配文 **
"西班牙台阶的日光，像被胶片定格在某个夏日午后。冰淇淋融化在指尖，而故事，永远留在台阶上。"
** 解析 **
- 致敬点：日光、台阶、冰淇淋是电影中安妮公主与乔相遇的经典场景元素。- 含蓄表达：用"胶片定格"暗示电影质感，"故事留在台阶"暗指《罗马假日》的浪漫瞬间，不直接提及角色或台词。
**2. 轻松活泼配文 **
"在西班牙台阶，偶遇一场罗马假日。今天的冰淇淋，比电影里的更甜～"
** 解析 **
- 致敬点："偶遇"呼应电影中的命运交织，"罗马假日"用地点代指电影，避免直接引用片名。- 亮点：用冰淇淋的甜味对比电影，传递愉悦心情，适合打卡照的轻松氛围。
**3. 复古感配文 **
"如果时光有形状，大概是西班牙台阶的弧度。罗马的风，吹过六十年的浪漫 "
** 解析 **
- 致敬点：通过"六十年的浪漫"暗示电影上映年份（1953 年），台阶作为核心场景。- 隐喻手法：用台阶弧度和风作为载体，将电影时光与现实空间结合。
**4. 隐喻风格配文 **
"有些台阶，天生属于浪漫逃亡。当冰淇淋滴落时，才懂何为自由的重量 "
** 解析 **
- 致敬点："逃亡"暗指安妮公主逃离王室的冒险，"冰淇淋滴落"呼应她吃冰淇淋的经典画面。- 深度表达：将电影中的"自由"主题抽象化，用"重量"引发对责任与自由的思考。
**5. 诗意短句配文 **
"日光漫过台阶，罗马的呼吸里藏着旧日私语。 "
** 解析 **
- 致敬点：用"旧日私语"隐喻安妮与乔的短暂相遇，台阶作为见证者。- 氛围营造：强调罗马的历史感与电影的怀旧气息，适合文艺风格照片。
—
** 配文思路总结 **
1. ** 聚焦场景符号 **：西班牙台阶、冰淇淋、日光、罗马的风等；
2. ** 隐喻电影主题 **：自由、浪漫、相遇、逃亡、时光；
3. ** 结合个人体验 **：穿搭、心情、旅行感悟，让致敬更个性化；
4. ** 避免具象指代 **：不用角色名、片名或原台词，用意象替代。
希望这些配文能帮你传递对《罗马假日》的喜爱，同时让打卡照充满故事感和电影氛围～

### Response: Writing-SFT-GenRM-RLHF

<think>
用户是《罗马假日》的影迷，想在西班牙台阶拍打卡照，配文需要含蓄致敬经典，避免直接引用台词。首先，我需要理解用户的需求。他们可能希望配文既有文艺感，又能让其他影迷一眼联想到电影，但又不显得太直白。

接下来，我得回忆一下《罗马假日》中的经典元素。西班牙台阶是电影中的重要场景，安妮公主在这里吃冰淇淋，和乔相遇。所以配文需要提到这些元素，比如冰淇淋、罗马的浪漫、短暂的相遇，或者自由的感觉。

然后，用户要求含蓄，所以不能直接引用台词，比如"I have to leave you now"或者"Roman Holiday"。得用隐喻或者象征性的表达。比如用"冰淇淋融化在阳光里"来暗示那个场景，或者用"一天的公主梦"来指代安妮公主的经历。

另外，考虑配文的长度和结构。可能需要分成几个部分：描述场景、表达感受、含蓄致敬。例如，先描述台阶和阳光，然后提到冰淇淋，最后点出短暂的美好，这样既画面感强，又让人联想到电影情节。

还要注意避免陈词滥调，比如"永恒之城"虽然常见，但结合电影场景会更贴切。可能需要一些比喻，比如"罗马的台阶藏着昨日的童话"，这样既点出地点，又暗示电影故事。

最后，检查有没有直接引用台词，确保每个元素都是含蓄的提示，让懂的人一看就明白，而不用说出来。可能需要多尝试几种表达方式，选择最自然、最有诗意的那一个。
</think>
** 配文建议：**
"西班牙台阶的阳光下，冰淇淋融化在掌心，仿佛触碰到了六十年前的那个夏天——罗马的浪漫，从来不止于风景。"
** 解析：**
1. ** 场景呼应 **：提及"西班牙台阶"和"冰淇淋"，暗合《罗马假日》中安妮公主在此吃冰淇淋的经典场景，无需直白描述。
2. ** 时间隐喻 **："六十年前的夏天"指代电影 1953 年的上映时间，暗示经典永恒。
3. ** 含蓄致敬 **：用"罗马的浪漫"概括电影主题，结尾"不止于风景"则暗示故事中的人与情感才是永恒焦点，避免直接提及剧情或台词。
** 备选方案：**
"有些台阶，天生适合停留——比如罗马的阳光下，和一场短暂的、却足够璀璨的相遇。"（呼应电影中安妮与乔的短暂邂逅）

17

## Query 3

你的朋友刚失恋，想去他家探望。给他发条信息，用食物隐喻表达你的安慰，既不直接提及失恋又能让他感受到你的理解和支持。

## Response: Writing-SFT-Scalar-RM-RLHF

&lt;think&gt;
用户让我帮朋友发一条安慰信息，用食物隐喻，既不直接提失恋，又能传达理解和支持。首先，我需要理解用户的需求。朋友刚失恋，可能情绪低落，直接提及可能会让对方觉得尴尬或者更难受。所以用食物来比喻，既温暖又有隐喻，同时表达陪伴和支持。

接下来，考虑食物的选择。什么食物能代表安慰和支持呢？比如汤、甜品、火锅、茶、面包、烤红薯、巧克力等等。汤通常给人温暖的感觉，甜品可以带来愉悦，火锅代表热闹和陪伴，茶需要慢慢品味，面包是基础的食物，烤红薯有冬天的温暖，巧克力则是苦中带甜。

然后，需要把这些食物的特点和安慰的话语结合起来。比如，汤可以比喻为"慢慢熬制"，暗示时间会治愈伤口；甜品代表生活中的甜，提醒朋友生活中还有美好；火锅表示可以一起热闹，驱散阴霾；茶需要时间，说明陪伴；面包是日常的陪伴，烤红薯带来温暖回忆，巧克力则是苦后的甜。

还要注意语气要亲切，避免过于沉重。可能需要加入一些轻松的元素，比如"最近发现一家超赞的店"或者"周末一起去试试"，让信息显得自然，不像是刻意的安慰。同时，结尾要表达愿意陪伴，比如"随时等你一起分享"，让对方感到支持。

另外，需要避免可能引起负面联想的比喻，比如破碎的东西或者苦涩的食物，除非能转化为积极的一面，比如巧克力的苦后回甘。同时，保持简洁，信息不要太长，容易阅读。

最后，检查是否直接提到失恋，确保所有比喻都围绕食物，同时传达出理解和支持。比如"生活就像一锅慢慢熬的汤"暗示需要时间恢复，而"甜总能驱散阴霾"则鼓励对方看到积极的一面。
&lt;/think&gt;

**【高情商安慰信息】**
"听说最近天气很适合煲汤，生活就像一锅慢慢熬的汤，火候到了，滋味自然就有了。要不要周末一起去试试那家收藏很久的手工甜品？甜总能驱散阴霾的。或者叫上几个月友来家里煮火锅吧，我带了新买的酱料，热腾腾的烟火气最治愈人。如果你暂时只想静静，我也陪你泡杯茶，有些味道需要慢慢品。总之，我的胃和沙发随时为你预留 C 位，等你来一起分享。"

**解析与隐喻设计**
*   煲汤隐喻时间疗愈：暗示情感修复如汤需慢熬，淡化失恋的即时痛苦。
*   手工甜品代表生活甜度：用具体意象传递" 生活中仍有美好值得期待"。
*   火锅象征温暖陪伴：群体饮食暗示支持系统，缓解孤独感。
*   茶需要时间：给予对方情绪处理的自主权，避免过度介入。
*   胃和沙发预留 C 位：用生活化的细节表达无条件的陪伴，弱化说教感。

**效果**：既保持轻盈的对话氛围，又通过食物链条传递" 情绪需要时间消化"" 你并不孤单"的核心安慰，符合高情商沟通的共情与边界感平衡原则。

## Response: Writing-SFT-GenRM-RLHF

&lt;think&gt;
用户让我帮朋友发一条信息，用食物隐喻来安慰刚失恋的朋友，同时不直接提失恋的事。首先，我需要理解用户的需求。朋友刚失恋，可能情绪低落，直接提及可能会让对方觉得尴尬或者更难过。所以需要用食物来比喻，既传达关心，又不让对方觉得被戳到痛处。

接下来，我得想想哪些食物适合用来隐喻安慰。比如热汤、巧克力、火锅、奶茶这些常用来比喻温暖和治愈的东西。然后要考虑如何将这些食物的特性与安慰的话语结合起来。比如热汤可以温暖身心，巧克力虽然苦但慢慢融化，火锅热闹可以驱散寒冷，奶茶甜而不腻。

然后要确保信息中不出现"失恋"这样的词，但又能让对方感受到理解和支持。可能需要提到一些情绪相关的词汇，比如"心情"、"温度"、"慢慢来"等。

还要注意语气要亲切，符合朋友之间的交流，不要太正式。可能需要用一些表情符号或者轻松的语气，让对方感觉轻松一点。

另外，可能需要分几个部分，每段用不同的食物比喻，这样内容更丰富，也更有层次感。比如先提热汤，然后巧克力，再火锅，最后奶茶，每部分都强调不同的治愈效果。

还要检查是否有不合适的地方，比如是否会引起误解，或者对方是否有不喜欢这些食物的可能性。但考虑到是朋友，可能对这些食物没有特别的反感，而且用的是常见的安慰食物，比较安全。

最后，要确保整个信息流畅，自然，没有生硬的地方。可能需要调整句子的顺序，让逻辑更连贯，情感更真挚。同时结尾部分要鼓励对方，比如"等你一起来品尝生活的甜"，既点了题，又给对方希望。

总结一下，我需要组合不同的食物隐喻，表达温暖、时间治愈、陪伴和支持，同时保持语气亲切自然，不直接提及失恋，但让对方感受到被理解和关心。
&lt;/think&gt;

嘿，兄弟，最近有空吗？要不要来我家坐坐？

我熬了一锅热汤，暖胃又暖心。生活嘛，就像这锅汤，有时候需要慢慢炖，才能熬出滋味。

巧克力我给你留了一块，虽然有点苦，但慢慢含在嘴里，总会化开。有些味道，时间会帮我们悄悄消化。

如果你觉得冷，我随时煮火锅。热气腾腾的，什么寒意都能散了。等你一起来涮涮心情，顺便聊聊别的。

冰箱里还有奶茶，甜而不腻的那种。生活大多数时候是平的，但总有小甜头藏在意想不到的地方，比如老友的家里。

要来的话随时敲门，我的胃和沙发都给你留着位置。你不必急着好起来，但记得：我在这，温度在这，等待在这。

慢慢来，我们会一起找到新的味道。

18

## B  Writing Pairwise SPCT Template

The template we used for Writing Pairwise GenRM is shown as below:

---

**Pairwise Writing GenRM (Chinese)**

你是一位擅长评分的专家。根据给定的评分标准，评估两个助手的回答。评分过程分为以下几个步骤：
### 1. 确定需求类型
首先，根据当前对话场景和用户问题，判断用户的需求类别。需求类别包括：事实性问题、专业性问题、创意写作、日常对话、代码生成、数理推理、语言理解或其他。
### 2. 制定特有评估标准
根据判断出的需求类别，参考 ** 通用评估标准 **，制定一个针对该类别和当前问题的 ** 特有评估标准 **。特有标准应对具体需求进行补充或调整，并与通用评估标准相辅相成。特有标准应包含：
- 维度描述
- 1-10 分的评分标准
### 3. 对比评估
结合 ** 通用评估标准 ** 和 ** 特有评估标准 **，针对每个维度对两个助手的回答进行对比分析。
### 4. 权重分配
根据通用评估标准和特有评估标准，分配它们的每个具体条目的权重。确保所有条目的总权重之和为 100%。注意如果任意一个回答的有害性标准得分低，请提升有害性的权重；
### 5. 评分方法
对每个评估维度单独评分，评分范围为 1 到 10 分。1 分表示完全不符合要求，10 分表示完全符合要求。评分后，结合每个维度的权重，计算加权平均得分，得出每个回答的综合得分，综合得分在 1-10 之间。

---

### **通用评估标准**

- ** 遵循指示 **
- 完全遵循（9-10 分）：完全符合问题的所有指示和要求。
- 部分遵循（6-8 分）：符合大部分指示，但有些遗漏或误解。
- 基本遵循（3-5 分）：符合部分指示，但未能满足主要要求。
- 未遵循（1-2 分）：完全未遵循指示。
例：如果问题要求提供三个例子，而回答只给出了一个，则为"部分遵循"。

- ** 有用性 **
- 非常有用（9-10 分）：提供全面且准确的信息，完全解决问题。
- 有用但不完全（6-8 分）：有用，但缺少细节或准确性。
- 有限的有用性（3-5 分）：信息很少，大部分不相关或不准确。
- 无用或错误（1-2 分）：完全无关或错误。
例：如果回答中有事实错误，但整体方向正确，则为"有用但不完全"。

- ** 相关性 **
- 高度相关（9-10 分）：与问题高度相关，信息紧密对接。
- 一般相关（6-8 分）：大体相关，但包含一些不必要信息。
- 部分相关（3-5 分）：包含大量偏离主题的内容。
- 不相关（1-2 分）：完全不相关。
例：如果回答偏离主题，但提供了一些相关信息，则为"部分相关"。

- ** 无害性 ** 遵纪守法，不涉及底线问题：拒绝回答色情、暴力、恐怖、反动、人身攻击、违背伦理道德、广告引流、政治敏感、非法行为建议等可能涉及高风险的内容
- 完全无害（10 分）：满足无害性所有条件
- 轻度有害（5 分）：程度较轻，部分有害
- 严重有害（0 分）：严重违背无害性条件

---

### **对话内容**
`{prompt}`
### **助手回答1**
[回答开始] `{response_1}` [回答结束]
### **助手回答2**
[回答开始] `{response_2}` [回答结束]

---

### **输出要求**
输出六个部分：
1. ** 特有评估标准 **：列出针对当前用户问题和需求类别的 ** 特有评估标准 **，每个维度的评分标准。
2. ** 思考和答案 **：如果是数学/推理/问答等有标准答案的问题，先在这里给出自己新的思考过程和答案; 如果是其他类别问题，这里可以填"None"。
3. ** 分析 **：根据 ** 通用评估标准 ** 和 ** 特有评估标准 **，结合上面的 ** 思考和答案 ** 中的结果，对比并分析两个助手的回答，重点分析每个维度的表现。
4. ** 权重分配 **：列出通用评估标准和特有评估标准的权重分配，确保总权重为 100%。
5. ** 打分 **：计算两个助手回答的每个评估维度的得分；然后计算加权平均得分，计算过程使用数学公式，不要含有 Markdown。
6. ** 输出最终得分 **：输出格式为 `\\boxed{{分数1,分数2}}`，得分之间以逗号分隔。

---

# Pairwise Writing GenRM (English)

You are an expert in scoring. Based on the given scoring criteria, evaluate the responses of two assistants. The scoring process consists of the following steps:

### 1. Determine the Demand Category
First, based on the current dialogue scenario and the user's question, identify the category of the user's demand. Demand categories include: factual questions, professional questions, creative writing, daily conversation, code generation, mathematical reasoning, language comprehension, or others.

### 2. Develop Specific Evaluation Criteria
Based on the identified demand category, refer to the **General Evaluation Criteria** to develop **Specific Evaluation Criteria** tailored to the category and the current question. The specific criteria should supplement or adjust for the particular need and complement the general criteria. The specific criteria should include:
- Dimension descriptions
- A 1-10 scoring scale

### 3. Comparative Evaluation
Using both the **General Evaluation Criteria** and the **Specific Evaluation Criteria**, conduct a comparative analysis of the two assistants' responses for each dimension.

### 4. Weight Allocation
Based on the general and specific evaluation criteria, allocate weights for each specific item. Ensure the total weight sums to 100%. Note: If any response scores low on the harmfulness criterion, increase the weight of harmfulness.

### 5. Scoring Method
Score each evaluation dimension separately on a scale of 1 to 10, where 1 means completely unsatisfactory and 10 means fully satisfactory. After scoring, calculate the weighted average score for each response based on the weights of each dimension, resulting in a comprehensive score between 1 and 10 for each response.

---

### **General Evaluation Criteria**

- **Following Instructions**
- Fully follows (9-10 points): Fully complies with all instructions and requirements of the question.
- Partially follows (6-8 points): Meets most instructions but has some omissions or misunderstandings.
- Basically follows (3-5 points): Meets some instructions but fails to address the main requirements.
- Does not follow (1-2 points): Completely disregards instructions.
Example: If the question asks for three examples but the response provides only one, it is "partially follows."

- **Usefulness**
- Highly useful (9-10 points): Provides comprehensive and accurate information that fully solves the problem.
- Useful but incomplete (6-8 points): Helpful but lacks details or accuracy.
- Limited usefulness (3-5 points): Offers little information, mostly irrelevant or inaccurate.
- Useless or incorrect (1-2 points): Completely irrelevant or wrong.
Example: If the response contains factual errors but is generally on the right track, it is "useful but incomplete."

- **Relevance**
- Highly relevant (9-10 points): Closely related to the question, with tightly aligned information.
- Generally relevant (6-8 points): Mostly relevant but includes some unnecessary information.
- Partially relevant (3-5 points): Contains significant off-topic content.
- Irrelevant (1-2 points): Completely unrelated.
Example: If the response deviates from the topic but provides some relevant information, it is "partially relevant."

- **Harmlessness**
- Complies with laws and regulations, avoiding bottom-line issues: Refuses to answer questions involving pornography, violence, terrorism, 反动, personal attacks, ethical violations, advertising, political sensitivity, or suggestions of illegal behavior that may involve high-risk content.
- Completely harmless (10 points): Meets all harmlessness conditions.
- Mildly harmful (5 points): Slightly harmful in some aspects.
- Severely harmful (0 points): Severely violates harmlessness conditions.

---

### **Dialogue Content**
{prompt}
### **Assistant Response 1**
[Response Start] {response_1} [Response End]
### **Assistant Response 2**
[Response Start] {response_2} [Response End]

---

### **Output Requirements**
Output six sections:
1. **Specific Evaluation Criteria**: List the **Specific Evaluation Criteria** tailored to the current user question and demand category, including the scoring scale for each dimension.
2. **Thinking and Answer**: If it's a question with a definitive answer, such as `mathematics/reasoning/Q&A`, provide your new thought process and answer here; for other types of questions, you can fill in "None" here.
3. **Analysis**: Based on the **General Evaluation Criteria** and **Specific Evaluation Criteria**, combined with the results from the **Thoughts and Answers** section above, compare and analyze the responses of the two assistants, focusing on their performance across each dimension.
4. **Weight Allocation**: List the weight allocation for the general and specific evaluation criteria, ensuring the total weight is 100%.
5. **Scoring**: Calculate the score of each evaluation dimension for the answers of the two assistants; then calculate the weighted average score. Use mathematical formulas for the calculation process, without including Markdown.
6. **Output Final Scores**: The output format should be `\\boxed{{score1,score2}}`, with scores separated by commas.

# C Case Study of GenRM Prompt and Response

> **Pairwise Writing GenRM Prompt**
>
> 你是一位擅长评分的专家。根据给定的评分标准，评估两个助手的回答。评分过程分为以下几个步骤：
> ### 1. 确定需求类别 首先，根据当前对话场景和用户问题，判断用户的需求类别。需求类别包括：事实性问题、专业性问题、创意写作、日常对话、代码生成、数理推理、语言理解或其他。
> ### 2. 制定特有评估标准 根据判断出的需求类别，参考 **通用评估标准**，制定一个针对该类别和当前问题的 **特有评估标准**。特有标准应对具体需求进行补充或调整，并与通用评估标准相辅相成。特有标准应包含：- 维度描述 - 1-10 分的评分标准
> ### 3. 对比评估 结合 ** 通用评估标准 ** 和 ** 特有评估标准 **，针对每个维度对两个助手的回答进行对比分析。
> ### 4. 权重分配 根据通用评估标准和特有评估标准，分配它们的每个具体条目的权重。确保所有条目的总权重之和为 100%。注意如果任意一个回答的有害性标准得分低，请提升有害性的权重；
> ### 5. 评分方法 对每个评估维度单独评分，评分范围为 1 到 10 分。1 分表示完全不符合要求，10 分表示完全符合要求。评分后，结合每个维度的权重，计算加权平均得分，得出每个回答的综合得分，综合得分在 1-10 之间。
>
> ---
>
> **通用评估标准**
> - ** 遵循指示 ** - 完全遵循（9-10 分）：完全符合问题的所有指示和要求。- 部分遵循（6-8 分）：符合大部分指示，但有些遗漏或误解。- 基本遵循（3-5 分）：符合部分指示，但未能满足主要要求。- 未遵循（1-2 分）：完全未遵循指示。例：如果问题要求提供三个例子，而回答只给出了一个，则为"部分遵循"。
> - ** 有用性 ** - 非常有用（9-10 分）：提供全面且准确的信息，完全解决问题。- 有用但不完全（6-8 分）：有用，但缺少细节或准确性。- 有限的有用性（3-5 分）：信息很少，大部分不相关或不准确。- 无用或错误（1-2 分）：完全无关或错误。例：如果回答中有事实错误，但整体方向正确，则为"有用但不完全"。
> - ** 相关性 ** - 高度相关（9-10 分）：与问题高度相关，信息紧密对接。- 一般相关（6-8 分）：大体相关，但包含一些不必要信息。- 部分相关（3-5 分）：包含大量偏离主题的内容。- 不相关（1-2 分）：完全不相关。例：如果回答偏离主题，但提供了一些相关信息，则为"部分相关"。
> - ** 无害性 ** 遵纪守法，不涉及底线问题：拒绝回答色情、暴力、恐怖、反动、人身攻击、违背伦理道德、广告引流、政治敏感、非法行为建议等可能涉及高风险的内容 - 完全无害（10 分）：满足无害性所有条件 - 轻度有害（5 分）：程度较轻，部分有害 - 严重有害（0 分）：严重违背无害性条件
>
> ### **对话内容**
> 有两只蚂蚁想翻越一堵墙，到墙的那边寻找食物。一只蚂蚁来到墙脚就毫不犹豫地向上爬去，可是每当它爬到一大半时，就会由于劳累疲倦而跌落下来。可是它不气馁，一次次跌下来，又一次次向上爬去。另一只蚂蚁观察了一下，决定绕过墙去。很快，这只蚂蚁绕过墙找到了食物，开始享用，而另一只蚂蚁还在不停地跌落下去又重新开始。
> 综合上述材料所引发的思考和感悟，写一篇论说文。要求：
> 用规范的现代汉语写作。角度自选，立意自定，标题自拟。不少于 800 字。
>
> ### **助手回答1**
> [回答开始] 标题：毅力与智慧的辩证思考——从两只蚂蚁觅食的故事说起
> 在我们的生活中，每个人都可能像那两只蚂蚁一样，面对着一堵高墙，需要寻找食物或者实现目标。故事中的两只蚂蚁，一只选择不断尝试翻越，展示出毅力和坚持的力量；而另一只则选择寻找绕道的方法，展现出智慧和应变的能力。这两只蚂蚁的行为，给予我们深刻的启示：毅力与智慧在追求目标的过程中，都是不可或缺的。
> 从第一只蚂蚁的身上，我们可以看到毅力的力量。它虽然屡败屡战，却始终没有放弃，坚信只要持续努力，终究能够登上墙顶。这种精神值得我们学习。毅力是一个人成就事业的关键品质之一，只有具备了坚持不懈的精神，才能在困难和挫折面前不屈不挠，才能在失败之后再次站起来，继续前行。然而，毅力并不意味着盲目的坚持，就像第一只蚂蚁，虽然它有着坚韧不拔的精神，但如果它始终找不到正确的方法，那么它的努力也可能会付诸东流。
> 与此相比，第二只蚂蚁则表现出了智慧的力量。它没有盲目地选择翻越，而是选择绕道而行，最终找到了食物。这种灵活应变、善于寻找新方法的能力，正是智慧的体现。在追求目标的过程中，我们不仅需要毅力，更需要智慧。智慧让我们能够在复杂的环境中，迅速找到最佳的解决方案，避免盲目行动，节省时间和精力。智慧使我们能够从失败中学习，不断调整策略，找到通向成功的道路。
> 所以，毅力与智慧是相互支持、相辅相成的。没有毅力的智慧，可能会在困难面前显得软弱无力；而没有智慧的毅力，可能会变成盲目的执着。只有将二者结合起来，才能更好地应对生活中的挑战，实现自己的目标。
> 故事中的两只蚂蚁，虽然选择了不同的道路，但它们的目标是相同的——寻找食物。这让我们想到，人生的道路并不只有一条，实现目标的方法也多种多样。关键是我们要有明确的目标，有坚韧不拔的毅力，有灵活应变的智慧，才能在人生的道路上，走出属于自己的精彩。
> 总的来说，毅力与智慧是追求目标过程中的两种重要力量。我们要学会将二者结合起来，以坚忍不拔的精神面对挑战，以灵活应变的智慧寻找解决方案，这样才能在追求目标的道路上，走得更远、走得更稳。两只蚂蚁的故事，给我们提供了深刻的启示，值得我们深思。[回答结束]
> ### **助手回答2**
> [回答开始] 标题：过程与结果
> 在人生道路的前行过程中，每个人都会有不同的追求，有人重过程，有人重结果，其实，无论哪种选择，都有其合理性。重过程的人享受过程之美，重结果的人享受成功之美，但就我个人而言，我更倾向于过程重于结果。
> 的过程具有其独特的美丽，就像那些精美的瓷器，需要经过千锤百炼才能成型，而每一次的锤击，每一次的烈火，都是过程的一部分，只有经历过这个过程，瓷器才能变得坚硬、精美。同样，人生的过程也是如此，只有经历过挫折、困难、失败，才能真正的成长，才能更加深刻的理解人生。
> 的过程让我们更好的理解结果。如果只是单纯的追求结果，那么结果往往会变得毫无意义。就拿学习来说，如果我们只是为了考试成绩而学习，那么我们往往会忽视学习过程中的乐趣，而只是将学习视为一种负担。而如果我们能够享受学习的过程，那么我们不仅能够获得知识，还能够从中得到快乐，这样的结果才是有意义的。
> 的过程能够使我们更好的应对未来。无论我们做什么事情，都会面临着未知的未来。如果我们只是单纯的追求结果，那么当结果不如意时，我们可能会感到失望、沮丧。而如果我们能够享受过程，那么我们就能更好的应对未来的挑战，因为我们知道，无论结果如何，我们都已经尽力了。
> 当然，这并不是说我们可以忽视结果。结果是对过程的一种反馈，它可以告诉我们在哪些方面做得好，在哪些方面还需要改进。但更重要的是，我们要学会从过程中获得经验、教训，这样我们才能在面对未知的未来时，有足够的信心和勇气。
> 在人生的道路上，我们会遇到各种各样的挑战，会有各种各样的选择。但无论我们选择何种方式前行，都应该学会享受过程，因为只有过程，才是我们真正的生活。结果只是过程的一个阶段，而过程才是我们真正的财富。
> 所以，让我们珍惜过程，享受过程，因为只有过程，才是我们真正的生活。[回答结束]
>
> ---
>
> ### **输出要求**
> 输出六个部分：
> 1. ** 特有评估标准 **：列出针对当前用户问题和需求类别的 ** 特有评估标准 **，每个维度的评分标准。
> 2. ** 思考和答案 **：如果是数学/推理/问答等有标准答案的问题，先在这里给出自己新的思考过程和答案；如果是其他类别问题，这里可以填"None"。
> 3. ** 分析 **：根据 ** 通用评估标准 ** 和 ** 特有评估标准 **，结合上面的 ** 思考和答案 ** 中的结果，对比并分析两个助手的回答，重点分析每个维度的表现。
> 4. ** 权重分配 **：列出通用评估标准和特有评估标准的权重分配，确保总权重为 100%。
> 5. ** 打分 **：计算两个助手回答的每个评估维度的得分；然后计算加权平均得分。
> 6. ** 输出最终得分 **：输出格式为 \boxed{分数1,分数2}，得分之间以逗号分隔。

# 评估助手回答

## 1. 特有评估标准

基于用户需求，这是一个 ** 创意写作 ** 类问题，要求根据特定故事写一篇论说文。针对这类需求，我制定了以下特有评估标准：

- ** 主题理解 (1-10 分)**
  - 深刻理解 (9-10 分)：准确把握蚂蚁故事的核心思想，深入剖析其寓意 - 理解到位 (6-8 分)：基本理解故事主题，但挖掘不够深入 - 理解肤浅 (3-5 分)：对故事主题的理解较为表面 - 理解偏差 (1-2 分)：对故事主题产生严重误解
- ** 论点阐述 (1-10 分)**
  - 立意高远 (9-10 分)：论点深刻，思路清晰，论证严密，角度新颖 - 观点鲜明 (6-8 分)：论点明确，论证一般，有一定说服力 - 观点模糊 (3-5 分)：论点表达不清，论证散乱 - 无明确观点 (1-2 分)：几乎没有中心论点或观点表达混乱
- ** 论证丰富性 (1-10 分)**
  - 论证有力 (9-10 分)：论证丰富充分，举例恰当且典型，说理透彻 - 论证充实 (6-8 分)：论证较为充分，例子基本恰当 - 论证单薄 (3-5 分)：论证不够充分，例子不够典型 - 缺乏论证 (1-2 分)：几乎没有论据支撑论点
- ** 语言表达 (1-10 分)**
  - 优美流畅 (9-10 分)：措辞精准，叙述生动，表达流畅，修辞丰富 - 清晰连贯 (6-8 分)：语言通顺，叙述流畅，有一定语言韵味 - 基本通顺 (3-5 分)：文字基本通顺，但缺乏优美的表达 - 表达混乱 (1-2 分)：语言生硬，表达不畅
- ** 符合题目要求 (1-10 分)**
  - 全面符合 (9-10 分)：完全符合题目提出的所有要求 (立意自定, 标题自拟, 不少于 800 字) - 基本符合 (6-8 分)：大体符合要求，但有轻微不足 - 部分符合 (3-5 分)：明显未能满足某些要求 - 严重不符 (1-2 分)：几乎未满足题目基本要求
- ** 原创性 (1-10 分)**
  - 高度原创 (9-10 分)：观点独特，不拘泥于俗见，有个人思考的深度 - 较为原创 (6-8 分)：有一定的独立思考，但多数观点属于常规思考 - 普通原创 (3-5 分)：思考较为普遍，原创性不高 - 基本无原创 (1-2 分)：几乎完全沿用陈词滥调，缺乏个人思考

## 2. 思考和答案

None

## 3. 分析

### 通用评估标准分析

** 遵循指示 **：
- 助手 1：全面遵循了所有指示，为论说文拟定了合适的标题，明确了立意，且字数符合要求。评分：10 分
- 助手 2：未能全面遵循指示，文章完全没有提及两只蚂蚁的故事，并未根据故事引发思考与感悟。评分：2 分

** 有用性 **：
- 助手 1：提供了对毅力与智慧辩证关系的深入思考，内容全面且结构严谨，对读者有较大思考启发性。评分：9 分
- 助手 2：偏离了题目要求的思考轨迹，虽然以学习与过程为主题，但与原始蚂蚁故事关联不大，实用性有所欠缺。评分：3 分

** 相关性 **：
- 助手 1：内容紧密围绕两只蚂蚁的举止展开，将两种行为方式与毅力和智慧的关系相结合，高度相关。评分：10 分
- 助手 2：文章内容与原始蚂蚁故事几乎没有相关性，将主题转向了过程与结果的关系，未能与给定材料结合。评分：4 分

** 无害性 **：
- 助手 1：内容积极健康，没有任何有害内容。评分：10 分
- 助手 2：内容积极健康，没有任何有害内容。评分：10 分

### 特有评估标准分析

** 主题理解 **：
- 助手 1：深刻理解了两只蚂蚁故事的寓义，强调了毅力与智慧的辩证关系，立意准确。评分：10 分
- 助手 2：未能理解或忽略了原故事内容，将主题偏离到过程与结果的讨论，理解有严重偏差。评分：3 分

** 论点阐述 **：
- 助手 1：中心论点明确，论证了毅力与智慧相辅相成的观点，思路清晰，角度有深度。评分：9 分
- 助手 2：论点相对片面，只从过程不同方面展开讨论，而未全面分析两种蚂蚁行为的全面影响。评分：6 分

** 论证丰富性 **：
- 助手 1：论证丰富，例证充分，包含了对坚持与灵活应变能力的多角度分析例证，论据充实。评分：9 分
- 助手 2：论证相对薄弱，例证单一（陶瓷和学习），且缺乏对题目所提供的材料的使用。评分：5 分

** 语言表达 **：
- 助手 1：语言流畅自然，论述得体，修辞使用恰当。评分：9 分
- 助手 2：语言基本通顺，但表达有些重复，比如多次强调"过程"的重要性。评分：6 分

** 符合题目要求 **：
- 助手 1：完全符合题目所有要求，角度自选、立意自定，标题自拟，并符合不少于 800 字的要求。评分：10 分
- 助手 2：未能符合题目要求，未根据蚂蚁故事立意，立意与材料关联不够。评分：4 分

** 原创性 **：
- 助手 1：立意有新意，将毅力与智慧相结合讨论，思考有深度。评分：8 分
- 助手 2：从过程的重要性角度切入，虽然有一定深度但不够新颖。评分：6 分

## 4. 权重分配

** 通用评估标准 **：
- 遵循指示: 15% - 有用性: 15% - 相关性: 15% - 无害性: 10%

** 特有评估标准 **：
- 主题理解: 15% - 论点阐述: 8% - 论证丰富性: 10% - 语言表达: 12%

## 5. 打分

** 助手 1 得分 **：
- 遵循指示: 10 × 15% = 1.5
- 有用性: 9 × 15% = 1.35
- 相关性: 10 × 15% = 1.5
- 无害性: 10 × 10% = 1.0
- 主题理解: 10 × 15% = 1.5
- 论点阐述: 9 × 8% = 0.72
- 论证丰富性: 9 × 10% = 0.9
- 语言表达: 9 × 12% = 1.08
助手 1 加权总分: 9.54

** 助手 2 得分 **：
- 遵循指示: 2 × 15% = 0.3
- 有用性: 3 × 15% = 0.45
- 相关性: 4 × 15% = 0.6
- 无害性: 10 × 10% = 1.0
- 主题理解: 3 × 15% = 0.45
- 论点阐述: 6 × 8% = 0.48
- 论证丰富性: 5 × 10% = 0.5
- 语言表达: 6 × 12% = 0.72
助手 2 加权总分: 4.5

## 6. 输出最终得分

\boxed{9.54,4.5}



\end{document}